\pdfoutput=1

\documentclass[11pt]{article}

\usepackage[final]{acl}

\usepackage{times}
\usepackage{latexsym}

\usepackage[T1]{fontenc}

\usepackage[utf8]{inputenc}

\usepackage{microtype}

\usepackage{inconsolata}

\usepackage{graphicx}

\usepackage{acronym}
\usepackage[skip=0pt]{caption}
\usepackage[inline]{enumitem}
\usepackage{booktabs}
\usepackage{makecell}
\usepackage{multirow}
\usepackage{subcaption}
\usepackage{adjustbox}
\usepackage[normalem]{ulem}
\useunder{\uline}{\ul}{}
\usepackage{xspace}
\usepackage{graphicx}
\usepackage{algorithm}
\usepackage{algorithmic}
\usepackage{tabularx}
\usepackage{tcolorbox}
\usepackage{color, xcolor}

\usepackage{amssymb} 
\usepackage{makecell} 
\usepackage{ulem}
\usepackage{listings}
\usepackage{xcolor}

\AtBeginDocument{%
  \providecommand\BibTeX{{%
    \normalfont B\kern-0.5em{\scshape i\kern-0.25em b}\kern-0.8em\TeX}}}

\newcommand{\ie}{\emph{i.e.,}\xspace}
\newcommand{\eg}{\emph{e.g.,}\xspace}
\newcommand{\eat}[1]{}
\newcommand{\groupOp}{\texttt{group}\xspace}
\newcommand{\rankOp}{\texttt{rank}\xspace}
\newcommand{\sumOp}{\texttt{summarize}\xspace}
\newcommand{\llmOp}{\texttt{LLM}\xspace}
\newcommand{\emphOp}[1]{{#1}\textsuperscript{\#}}

\newcommand{\seriaArg}[1]{\texttt{MSP}(#1)}
\newcommand{\sllmArg}[1]{\texttt{LLM}_s(#1)}
\newcommand{\pllmArg}[1]{\texttt{LLM}_p(#1)}

\newcommand{\stitle}[1]{\textbf{#1}.}

\newcommand{\ours}{InsightTab\xspace}
\newcommand{\summaryboost}{SumBoost\xspace}
\newcommand{\baseline}{Original\xspace}
\newcommand{\tablet}{Tablet\xspace}
\newcommand{\tabllm}{TabLLM\xspace}
\newcommand{\xgb}{XGBoost\xspace}
\newcommand{\tabpfn}{TabPFN\xspace}

\title{From Data to Insights: Integrating Data Modeling with LLMs for Tabular Classification}

\title{Insight Distillation Helps LLMs to Be Effective Tabular Classifiers}

\title{From Data to Insights: Integrating Data Modeling with LLMs for Tabular Classification}

\title{Summarize-Examplify-Reflect: Transforming Data into Insights for LLM-based Tabular Classification}

\title{Summarize-Exemplify-Reflect: Data-driven Insight Distillation Empowers LLMs for Few-shot Tabular Classification}

\author{Yifei Yuan$^{1,2}$, Jiatong Li$^{3}$, Weijia Zhang$^{4}$, Mohammad Aliannejadi$^{4}$, \\ \textbf{Evangelos Kanoulas}$^{4}$, \textbf{Renjun Hu}$^{5}$\thanks{corresponding author}\\
  $^1$ETH Zürich, $^2$University of Copenhagen, $^3$University of Science and Technology of China, \\$^4$ University of Amsterdam, $^5$ East China Normal University \\
  \texttt{yuanyif@ethz.ch}, \texttt{rjhu@dase.ecnu.edu.cn} }

\begin{document}
\maketitle

\begin{abstract}
Recent studies show the promise of large language models (LLMs) for few-shot tabular classification but highlight challenges due to the variability in structured data. To address this, we propose distilling data into actionable insights to enable robust and effective tabular classification by LLMs. Inspired from human learning processes, we introduce \ours, an insight distillation framework guided by principles of divide-and-conquer, easy-first, and reflective learning. It integrates rule summarization, strategic exemplification, and insight reflection through deep collaboration between LLMs and data modeling techniques. The obtained insights enable LLMs to better align their general knowledge and capabilities with the particular requirements of specific tabular tasks. We extensively evaluate \ours on nine datasets. The results demonstrate consistent improvement over state-of-the-art methods. Ablation studies further validate the principle-guided distillation process, and in-depth analyses emphasize \ours's effectiveness in leveraging labeled data and managing biases.
\end{abstract}

\maketitle

\acresetall

\section{Introduction}
Tabular data is widely used across applications due to the prevalence of relational databases~\cite{ShwartzZiv2021TabularDD}, inspiring tasks like tabular question answering~\cite{Vakulenko2017TableQAQA}, retrieval~\cite{Zheng2023DenseRL}, etc.
Among these tasks, tabular classification plays a key role by categorizing tabular data with numerical and categorical features into predefined classes. 

Traditional tabular classification tasks leverage ensemble techniques with decision trees such as XGBoost to handle mixed numerical and categorical data~\cite{ShwartzZiv2021TabularDD}. The high cost of labeling large datasets and feature engineering has driven interest in few-shot learning, where models make accurate predictions using only a few examples~\cite{Hegselmann2022TabLLMFC}. These approaches often leverage large language models (LLMs), which enhance prediction accuracy by transforming structured data into natural language sequences, enabling LLMs to process and make informed predictions. They show that combining proper serialization with LLMs achieves strong performance, with later studies refining this through expert and algorithmic prompting~\cite{Slack2023TABLETLF,Manikandan2023LanguageMA}.

However, several challenges persist for effective LLM-based tabular classification  in the few-shot setting. These include:
\begin{enumerate*}[label=(\arabic*)]
    \item \emph{Filling knowledge gap}. Tabular data often includes task-specific details (e.g., task description), failing to incorporate them can hinder the performance of general-purpose LLMs.
    \item \emph{Unlocking LLMs' full potential}. Effective tabular classification requires complex processing. While methods like in-context learning help, they fall short of fully utilizing LLMs' capabilities, making this an ongoing challenge.
    \item \emph{Balancing performance and cost}. Tabular classification tasks are critical for real-time services, demanding fast response times and cost efficiency. Traditional methods, while effective, are often time-intensive to train. Thus, LLM-based solutions must be both efficient and economical to meet these demands.
    
\end{enumerate*}

To address these challenges, we focus on the task of \textbf{LLM-based tabular classification in the few-shot setting}. We pose a fundamental research question: \textit{Can actionable insights be distilled from the training data to enhance LLMs’ performance in this task?} In response, we introduce \ours{}, a novel insight distillation framework that integrates traditional machine learning techniques with LLMs, enabling them to acquire task-specific knowledge from a data modeling perspective. In this analogy, LLMs can be seen as students with broad general knowledge but limited task-specific expertise.
Inspired by human learning processes, \ours employs three key principles: \textit{divide-and-conquer}, \textit{easy-first}, and \textit{reflective learning}. The \textit{divide-and-conquer} strategy, inspired by curriculum learning~\cite{Bengio2009CurriculumL}, partitions training data into subsets, enabling LLMs to derive rules for each, \ie a cognitive process akin to \textbf{summarization}. This could simultaneously avoid context window overflow and yield more targeted rules than directly feeding all data into an LLM. The \textit{easy-first} principle mimics the human cognitive procedure of \textbf{exemplifying}, prioritizing the selection of representative and simpler samples for initial learning~\cite{Sun2024EasytoHardGS}. Finally, the \textit{reflective learning} strategy allows LLMs to engage in \textbf{reflection}, \ie learning from misclassified cases to enhance their distilled insights.

We then use three operators for the insight distillation process: \verb|group| for clustering similar samples, \verb|rank| for example ordering with prediction difficulty, and \verb|summarize| for extracting natural language rules. This process fosters collaboration between data modeling methods and LLMs, leveraging LLMs' strength in summarization while addressing their limitation in data analysis.

We validate \ours through experiments on nine diverse datasets, demonstrating its superiority over state-of-the-art methods in average performance. Additionally, \ours outperforms other approaches using the same base LLMs in 19 out of 20 comparisons. This highlights the importance of insight distillation for few-shot tabular classification. Ablation and case studies further confirm the effectiveness of each module, and tests on biased data show \ours's priority in robustness, reducing overfitting and improving generalization. In summary, our contributions are threefold:
\begin{itemize}
    \item Conceptually,  we highlight the importance of insight distillation through a deep integration of data modeling methods and LLMs prompting for few-shot tabular classification. 
    \item Methodologically, we present \ours{}, an insight distillation framework inspired by human learning principles of divide-and-conquer, easy-first, and reflective learning.
    \item Experimentally, we validate the enhanced effectiveness of \ours{} through extensive tests, as well as its rationales and superiority in sample efficiency and robustness\footnote{Code and dataset available at \url{https://github.com/yfyuan01/InsightTab/}.}.
\end{itemize}

\begin{figure*}[t]
    \centering
    \includegraphics[width=\linewidth]{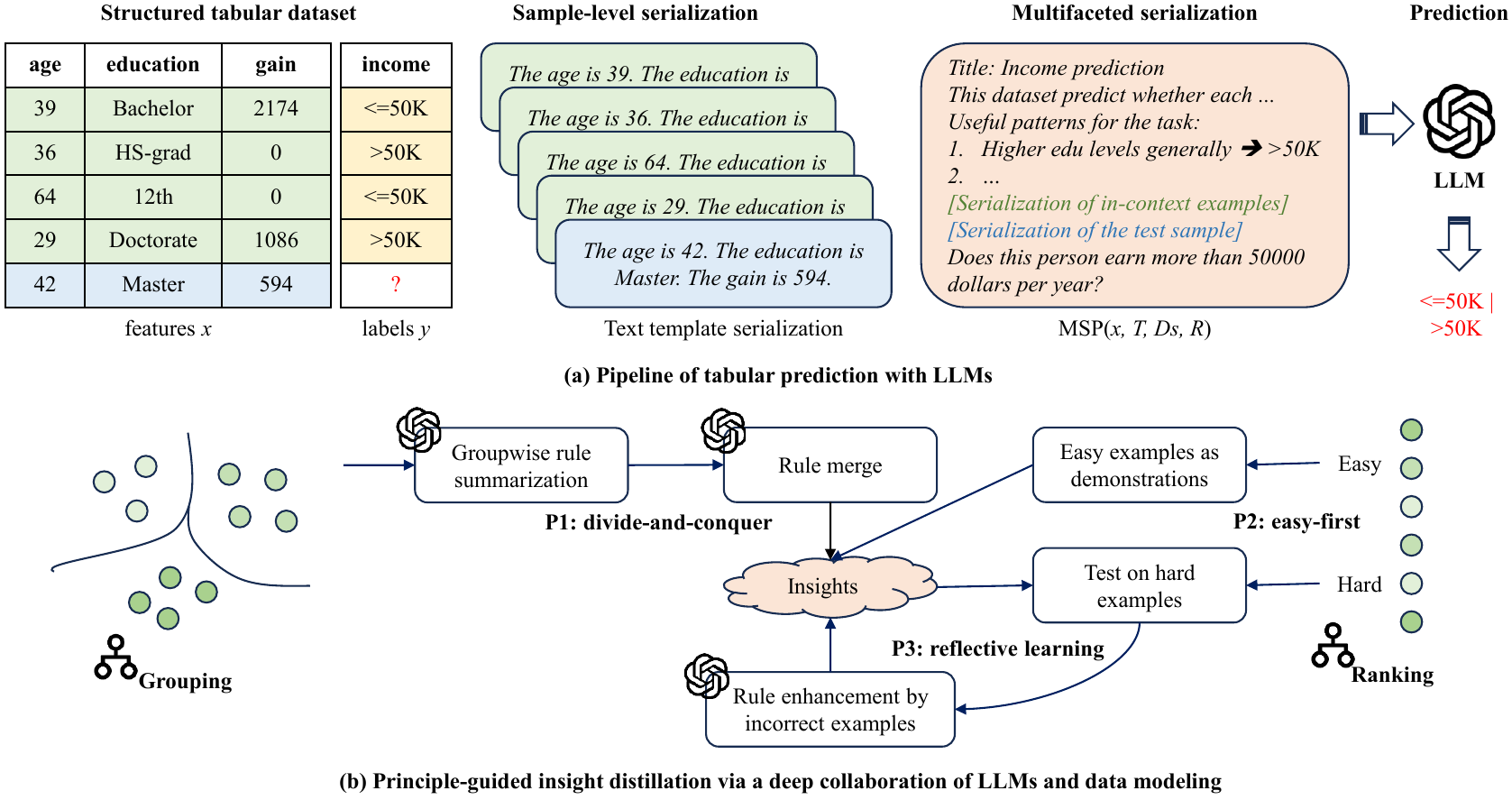}
    \caption{Framework overview of \ours. (a) We present an income prediction dataset with $n=4$ samples and $d=3$ features.
    The text template serialization converts column features into natural language sequences, while the multifaceted serialization further integrates them with task-specific knowledge into a prompt, for LLMs to predict test data. (b) Insight distillation involves both data modeling techniques and LLMs for sample grouping/ranking and rule summarization, respectively. The process is guided by three principles inspired from human learning.}
    \label{fig:model-structure}
\end{figure*}


\section{Related Work}

\stitle{Tabular classification}
Tabular classification assigns labels to table rows based on column attributes~\cite{Fang2024LargeLM}. Traditional methods use gradient-boosted decision trees~\cite{Chen2016XGBoostAS,Ke2017LightGBMAH} for structured data~\cite{ShwartzZiv2021TabularDD,Gorishniy2021RevisitingDL}, while others design specialized architectures~\cite{Chen2023TabCapsAC,Shah2022EnhancedTA,Du2021TabularNetAN}, such as TabularNet~\cite{Du2021TabularNetAN}, which encodes spatial and relational information.
Transformer-based approaches have also been explored~\cite{Hollmann2022TabPFNAT,Nassar2022TableFormerTS}, including models  specifically optimized for small tabular datasets like TabPFN~\cite{Hollmann2022TabPFNAT}. Recently, LLMs have gained attention for their few-shot learning abilities~\cite{Hu2022InContextLF}, enabling classification with minimal samples~\cite{Hegselmann2022TabLLMFC,Nam2023STUNTFT,Han2024LargeLM}. For instance, \citet{Nam2023STUNTFT} and \citet{qu2025tabicl} incorporate meta-learning and curriculum learning into LLM prompting, drawing inspiration from traditional data mining techniques.
Based on that, our approach 
focuses on LLM-based few-shot tabular classification and propose to distill actionable insights from tabular data. 
\stitle{LLMs for tabular data}
Language models have demonstrated effective capabilities in handling structured data~\cite{Lu2024LargeLM}. Extensive studies have explored pretraining language models to effectively learn structured table representations, ranging from smaller-scale pretrained models (\eg BERT, GPT-2)~\cite{Yin2020TaBERTPF,Liu2021TAPEXTP,Iida2021TABBIEPR} to state-of-the-art LLMs~\cite{Zha2023TableGPTTU,Zhang2023TableLlamaTO,Zhang2024TableLLMET}. For example,  ~\citet{Zhang2023TableLlamaTO} fine-tune Llama 2 to develop open-source LLMs capable of handling a variety of table-based tasks.
Since LLMs are designed for natural language, converting tabular data into text enables them to analyze and generate insights effectively~\cite{Nan2023EnhancingFT}. This process, known as data serialization, typically involves linearizing tables row by row with column separators~\cite{Sui2023TAP4LLMTP,Singha2023TabularRN}. For example,~\citet{Gao2023TexttoSQLEB} find that creating in-context learning prompts which include instruction helps improve Text-to-SQL performance.~\citet{Manikandan2023LanguageMA} are the first to generate summaries and shows that incorporating task-specific knowledge 
during serialization can improve tabular classification. Building on their findings, our work aims to develop a multifaceted serialization prompt that combines examples and rules from the training samples in natural language format, providing a specialized form of serialization. This approach provides strong guidance for LLMs, helping them understand the data structure.
\section{Methodology}
\label{sec:method}


In this section, we describe the structure of \ours{}, which integrates data modeling techniques with LLMs for better tabular classification. An overview of our approach is presented in Figure~\ref{fig:model-structure}. 



\subsection{Preliminaries}

\stitle{Tabular Data}
%
A labeled tabular dataset with $n$ rows (\ie samples) and $d$ columns (\ie features) could be denoted as $\mathcal{D} =\{(x^{(i)}, y^{(i)})\}_{i=1}^n$, where each $x^{(i)}$ is a $d$-dimensional feature vector and $y^{(i)} \in \mathcal{C}$ is the corresponding class label of $x^{(i)}$. $\mathcal{C}$ denotes the set of labels for a tabular classification task $T$. Each column of $\mathcal{D}$ is associated with a semantically meaningful feature name such as age, education, or marital status. These feature names are formally denoted as $\mathcal{F}=\{f_1, \dots, f_d\}$. 

\noindent
\stitle{Multifaceted Serialization}
A prerequisite for using LLMs in tabular classification is data serialization, which involves converting raw tabular data into natural language to enhance LLM understanding.
~\citet{Hegselmann2022TabLLMFC} empirically examined nine different serialization formats for tabular data and found that the \emph{Text Template} approach performs well across their experiments. Specifically, this approach converts a feature vector $x$ to a textual enumeration of all features as ``The $f_1$ is $x_{1}$. ... The $f_d$ is $x_{d}$.'', where $x_{j}$ represents the $j$-th value of $x$ (see Figure~\ref{fig:model-structure}). We adopt it as the default sample-level serialization method.
Additionally, concurrent studies~\cite{Slack2023TABLETLF,Manikandan2023LanguageMA} have verified that LLMs benefit from task-specific knowledge for tabular classification. The knowledge includes but not limited to task background, labeled few-shot examples, and rules mined from data. We hence extend the scope of serialization and define a multifaceted serialization prompt $\seriaArg{x, T, \mathcal{D}_s, R}$ which converts not only test sample $x$ but also the task $T$, few-shot in-context learning examples $\mathcal{D}_s \subset \mathcal{D}$, and mined rules $R$ to natural language prompt. Figure~\ref{fig:model-structure} also demonstrates a brief example of it (prompts are listed in Appendix \ref{appendix:prompts}). 

\noindent
\stitle{Task Formalization} 
%
With the aforementioned concepts, we formalize the LLM-based tabular classification task, particularly in few-shot settings. Given the tabular classification task $T$, an associated labeled dataset $\mathcal{D}$, and an unlabeled test sample $\tilde{x}$, the task goal is to predict a class label $\tilde{y} \in \mathcal{C}$ for $\tilde{x}$ with an LLM. This can be expressed as $\tilde{y} = \pllmArg{\seriaArg{\tilde{x},T,\mathcal{D}_s,R}}$, where $R$ represents the rules mined by the LLM from the training samples. 
These few-shot settings often occur in realistic scenarios where the labeled data is limited and has complex patterns. In these cases, LLMs can effectively leverage their rich language knowledge to make accurate predictions, whereas traditional methods struggle to fit an effective decision function with limited data. 
%



\subsection{Collaborative Insight Distillation} 

We propose a method \ours{} that enables LLMs to approach the task from a data modeling perspective and distill training data into actionable \emph{insights}. 
These insights are injected into the multifaceted serialization function of \ours{}, enabling LLMs to better adapt their general natural language understanding capabilities to different tabular prediction tasks.



In this context, the role of LLMs could be an analogy to students with extensive general knowledge but limited expertise in the specific task. 
As illustrated in Figure~\ref{fig:model-structure}, drawing from human learning strategies, we employ three key principles, namely \emph{divide-and-conquer}, \emph{easy-first}, and \emph{reflective learning}, to optimize the learning process for LLMs.
Specifically, we adopt a divide-and-conquer strategy for rule summarization: training samples are firstly grouped into subsets of `similar' examples. Summarizing rules for each subset individually then becomes more manageable than for the entire dataset. 
%
Next, following the easy-first principle, we choose the most confident samples as in-context demonstrations~\cite{dong2024surveyincontextlearning}. This allows LLMs to quickly acquire skills when confronting a new tabular classification task. 
Finally, we design a reflective learning mechanism for LLMs to refine their formed knowledge. In this mechanism, a subset of challenging examples is selected from the training data. LLMs are required to make predictions based on the existing rules and demonstrations. Incorrectly predicted samples are retained for further rule summarization, resulting in new rules that augment the existing ones. 

Overall, principle-guided insight distillation allows LLMs to draw task-specific knowledge from data, which ultimately enhances the overall tabular classification. To facilitate this insight distillation process, we employ three
key operators:
\begin{itemize}
    \item \groupOp divides samples into subsets with high intra-group similarity;
    \item \rankOp orders samples based on the difficulty for prediction;
    \item \sumOp derives a concise set of natural language rules from the samples.
\end{itemize}
In summary, while LLMs excel at \sumOp, they may fall short in the \groupOp and \rankOp operations which involve complex mathematical calculations, reasoning, and iterative refinement—tasks that are difficult to articulate clearly with natural language instructions. 
Conversely, \groupOp and \rankOp correspond to well-defined data modeling tasks that traditional machine learning methods handle effectively. Therefore, \ours{} forsters a  collaboration between LLMs and data modeling, enabling effective insight distillation for LLMs. Details are introduced in the next section.



\begin{table*}[!h]
\footnotesize
\centering
\begin{tabular}{cc c c c c c c c c c c}
\toprule
Method & $\llmOp_p$ & Bank & Blood & Calhou. & Car & Creditg & Diabe. & Heart & Income & Jungle & Avg \\
\midrule
 \xgb & / & 20.9 & 32.9 & 70.6 & 40.2 & 78.4 & 50.1 & 77.9 & 33.9 & \underline{67.9} & 52.5 \\
 \tabpfn & / & 14.3 & 17.6 & \textbf{77.8} & 30.5 & 81.3 & 46.7 & \textbf{81.8} & 20.2 & 65.3 & 48.4 \\ 
\midrule 
 \baseline & mistral-7b & 17.0 & 23.9 & 37.2 & 20.2 & 70.9 & 36.7 & 30.6 & 13.9 & 23.8 & 30.5 \\
 
 \tablet & mistral-7b & 10.8 & 29.8 & 54.9 & 25.4 & 66.7 & 39.8 & 42.3 & 24.0 & 39.0 & 37.0 \\
  
 \tabllm & mistral-7b & 21.7 & 38.0 & 69.2 & 35.3 & 80.9 & 59.2 & 74.1 & 58.5 & 63.8 & 55.6 \\

 \summaryboost & mistral-7b & 17.1 & 32.8 & 23.2 & 22.8 & 64.1 & 38.0 & 62.6 & 24.9 & 37.4 & 35.9 \\

 \ours & mistral-7b & \textbf{41.1}* & \textbf{51.8}* & \underline{77.2}*& \underline{48.6*} & \underline{81.9}* & \textbf{69.6}* & 76.7* & \textbf{66.5}* & \textbf{69.4}* & \textbf{64.8}* \\

 \midrule
 \baseline & gpt-3.5 & 28.9 & 45.4 & 68.2 & 32.9 & 78.2 & 54.1 & 58.7 & 44.9 & 54.6 & 51.8 \\
 
 \tablet & gpt-3.5 & 31.2 & 43.7 & 63.1 & 34.5 & 74.2 & 61.1 & 76.6 & 55.5 & 62.8 & 55.9 \\

 \tabllm & gpt-3.5 & 34.6 & 44.8 & 69.1 & 46.3 & \textbf{82.0}* & 61.3 & 79.5 & 59.9 & 62.5 & 60.0 \\
 
 \summaryboost & gpt-3.5  & 29.9 & 43.6 & 50.1 & 45.6 & 64.7 & 61.6 & 63.7 & 58.3 & 59.2 & 53.0 \\
 
 \ours & gpt-3.5 & \underline{38.8}* & \underline{50.7}* & 72.3* & \textbf{50.8}* & 81.4 & \underline{65.1}* & \underline{80.1}* & \underline{61.8}* & 64.4* & \underline{62.8}* \\ 
\bottomrule
\end{tabular}
\caption{Overall performance for few-shot tabular classification. We use 16/32/64/128 samples for training and 16 shots for in-context demonstration. The averaged F1 scores (\%)  are reported, \ie each number (except for the Avg column) is averaged from $5 \times 4=20$ tests. The best and second-best results are \textbf{bolded} and \underline{underlined}. * indicates the best performance among all methods using the same base LLM. Detailed results see Appendix~\ref{app:extra-exp-overall} and \ref{app:extra-exp-fulltrain}.}
\label{tab:overall}
\end{table*}

\subsection{Algorithm}
\label{subsec:algorithm}
\begin{algorithm}[t!]
\caption{\ours}
\label{alg:main}
\small
\begin{algorithmic}[1]
\REQUIRE classification task $T$, training data $\mathcal{D}$, test sample $\tilde{x}$, number $n_e$/$n_h$ of easy/hard samples, $\llmOp_s$ for rule summarization, $\llmOp_p$ for prediction
\ENSURE classification label $\tilde{y} \in \mathcal{C}$ of $\tilde{x}$
\STATE $M \leftarrow \texttt{XGB}(\mathcal{D})$, $M_1 \leftarrow$ the first tree in $M$
\STATE $F \leftarrow$ the number of leaves in $M_1$
\FOR{$i = 1$ to $F$}
    \STATE \groupOp: $\mathcal{D}_i \leftarrow$ samples in the $i$-th leaf of $M_1$
    \STATE \sumOp: $R_i \leftarrow \sllmArg{\mathcal{D}_i}$
\ENDFOR
\STATE Merge rules: $R \leftarrow \sllmArg{R_1, \dots, R_F}$
\STATE Compute: $h^{(i)} \leftarrow \texttt{entropy}(M(x^{(i)}))$, $i\in \{1, \dots, n\}$
\STATE \rankOp: $\mathcal{D}_{e} \leftarrow$ subset of $\mathcal{D}$ with the top-$n_e$ lowest $h^{(i)}$
\STATE \rankOp: $\mathcal{D}_{h} \leftarrow$ subset of $\mathcal{D}$ with the top-$n_h$ highest $h^{(i)}$
\STATE  $\mathcal{D}'_{h} \subset \mathcal{D}_{h} = \{(x, y)\}$ \emph{s.t.} $\pllmArg{\seriaArg{x, T, \mathcal{D}_{e}, R}} \ne y$
\STATE \sumOp: $R_h \leftarrow \sllmArg{\mathcal{D}'_{h}}$
\STATE Enhance rule: $R^+ \leftarrow R \oplus R_h$
\STATE Predict: $\tilde{y} \leftarrow \pllmArg{\seriaArg{\tilde{x}, T, \mathcal{D}_{e}, R^+}}$
\end{algorithmic}
\end{algorithm}

\noindent We introduce an algorithmic implementation of our method in Algorithm~\ref{alg:main}. In detail, we adopt gradient-boosting decision trees, \eg XGBoost~\cite{Chen2016XGBoostAS}, as the data modeling technique with the flexibility to adopt other approaches as long as the requirements of \groupOp and \rankOp are satisfied (we make a performance comparison in Appendix~\ref{appendix:comparative}).
Our algorithm takes the classification task description $T$, the training dataset $\mathcal{D}$, a test sample $\tilde{x}$, a summarizer LLM, and a predictor LLM as input, and outputs the corresponding label $\tilde{y}$ for $\tilde{x}$.
It first fits a XGBoost model $M$ on $\mathcal{D}$ to support \groupOp and \rankOp (lines 1--2). Specifically, we use the first tree of $M$ for sample grouping: samples within each leaf are grouped together and fed into a strong LLM $\llmOp_s$ for rule summarization (lines 3--6). Since the group-wise rules could contain redundancy, we then perform a rule merging process, resulting in a concise set of task-level rules (line 7). 
The method then computes the entropy of the predicted class probabilities $M(x^{(i)}) \in \mathbb{R}^{|\mathcal{C}|}$ of each training sample $x^{(i)} \in \mathcal{D}$ by the obtained model $M$ (line 8). These entropy values are used as the ranking scores for selecting easy and hard samples (lines 9--10). 
For each chosen hard sample, we let the predictor LLM $\llmOp_p$ (could be LLMs with a smaller size to save budget) make an intermediate prediction using easy examples as few-shot demonstrations and current rules as insights. Those incorrectly predicted hard samples are retained and sent to the summarizer LLM for rule enhancement (lines 11--13).  
Finally, we use the easy examples and updated rules as insights and derive the class prediction for $\tilde{y}$. An illustration of the mined rules see Appendix \ref{mined-rules}.

\section{Experiments}

\subsection{Experimental Setup}



\stitle{Datasets}
Following previous works~\cite{Fang2024LargeLM}, we evaluate our approach with nine datasets, all collected by~\cite{Hegselmann2022TabLLMFC}. These datasets, with their diverse sizes and feature sets, serve as an ideal testbed for LLM-based tabular classification methods (details in Appendix \ref{appendix:data}).


\stitle{Baselines} 
We select a broad spectrum of methods for comparison, with an emphasis on few-shot approaches. These include classic tree-ensemble model \textbf{\xgb}~\cite{Chen2016XGBoostAS}, pre-trained Transformer \textbf{\tabpfn}~\cite{Hollmann2022TabPFNAT}, a base LLM classifier \textbf{\baseline}, and more advanced LLM methods including \textbf{\tablet}~\cite{Slack2023TABLETLF}, \textbf{\tabllm}~\cite{Hegselmann2022TabLLMFC}, and \textbf{\summaryboost}~\cite{Manikandan2023LanguageMA}.

\stitle{Implementation} 
We employ both open- and close-sourced LLMs for prediction, \ie \textit{mistral-7b}~\cite{Jin2023TabCoTZT} and \textit{gpt-3.5-turbo}~\cite{Brown2020LanguageMA}. 
For rule summarization in the preprocessing step, we use \textit{gpt-4-turbo}. All base prompts are borrowed from \tabllm. We conduct five-fold cross-validation by randomly sampling some instances from the training split as the training set, and reporting the average F1 score for comparison. 

Detailed experimental setup information can be found in Appendix~\ref{appendix:exp} (datasets in \ref{appendix:data}, baselines in \ref{appendix:baselines}, and the parameter used in \ref{appendix:parameters}), with all prompt templates provided in Appendix~\ref{appendix:prompts}.


\begin{table*}[htp]
\centering
\small
\begin{tabular}{cc ccccccccc c}
\toprule
$\llmOp_p$ & Variant & Bank & Blood & Calhou. & Car & Creditg & Diabe. & Heart & Income & Jungle & Avg \\
\midrule
\multirow{4}{*}{mistral-7b} & - demostr. & 17.6 & \underline{48.1} & 13.9 & \emphOp{20.6} & \emphOp{11.5} & 51.8 & \emphOp{11.4} & \emphOp{31.0} & 46.3 & \emphOp{28.0} \\
& - grouping & \emphOp{10.1} & \emphOp{19.4} & \emphOp{12.3} & 44.3 & 12.9 & \emphOp{51.3} & 55.6 & 44.5 & \emphOp{37.8} & 32.0 \\
& - reflection & \underline{21.5} & 44.8 & \underline{62.1} & \underline{47.4} & \underline{36.6} & \underline{68.6} & \underline{56.2} & \underline{47.8} & \underline{48.9} & \underline{48.2} \\
& InsightTab & \textbf{35.3} & \textbf{50.6} & \textbf{78.2} & \textbf{55.0} & \textbf{82.9} & \textbf{70.8} & \textbf{74.2} & \textbf{63.4} & \textbf{66.3} & \textbf{64.1} \\ \midrule 
\multirow{4}{*}{\makecell{gpt-3.5 \\ -turbo}} & - demostr. & \emphOp{21.5} & \emphOp{40.9} & \emphOp{47.2} & \emphOp{36.4} & 47.3 & \emphOp{55.3} & \emphOp{15.8} & \emphOp{33.2} & \underline{62.4} & \emphOp{40.0} \\
& - grouping & 25.1 & \underline{43.0} & \underline{68.7} & 38.6 & \underline{57.1} & 59.6 & 64.0 & \textbf{59.5} & \emphOp{58.6} & \underline{52.7} \\
&- reflection & \underline{28.1} & 42.2 & 68.0 & \underline{46.6} & \emphOp{39.7} & \underline{60.6} & \underline{66.1} & 57.6 & 60.2 & 52.1 \\
& InsightTab & \textbf{37.9} & \textbf{50.2} & \textbf{72.6} & \textbf{50.4} & \textbf{79.8} & \textbf{64.2} & \textbf{83.}6 & \textbf{59.5} & \textbf{65.6} & \textbf{62.6} \\
\bottomrule
\end{tabular}
\caption{Ablation study results with 128 training samples, 16 shots, and averaged F1 scores (\%). The best, second-best and worst results within each group are in \textbf{bold}, \underline{underlined}, and with \emphOp{}, respectively.}
\label{tab:abalationstudy}
\end{table*}

\subsection{Overall Performance}


We present the overall performance in Table~\ref{tab:overall}. Our findings include: traditional tabular classification methods like \xgb rely heavily on large datasets for optimal performance, making them notably ineffective in few-shot scenarios, as seen with the Bank, Blood, and Income datasets. \tabpfn tackles the few-shot setting using prior-data modeling. However, it faces difficulties with class imbalance in certain datasets (\eg Bank and Blood).

The results affirm the potential of LLM-based methods for this task (\eg \tabllm). 
Building on this, \ours integrates data-driven insights with multifaceted serialization, leveraging summarization and reflection to outperform other LLM-based methods on \textbf{eight of nine} datasets. For the two datasets where \ours doesn't reach the best performance, it ranks the second and closely follows the non-LLM leaders. Overall, it demonstrates average improvements of 21\%, 32\%, 55\%, 37\%, 10\%, and 44\% over \xgb, \tabpfn, \baseline, \tablet, \tabllm, and \summaryboost, respectively. Interestingly, we find mistral-7b outperforms gpt-3.5 in \ours. This is because its architecture aligns better with structured, iterative reasoning tasks, whereas gpt-3.5 may overcomplicate or struggle with such setups. These findings highlight the advantage of LLMs with data-driven insight distillation for  robust few-shot tabular classification.

\subsection{Ablation Study}
We conduct an ablation study by selectively removing the easy example demonstration, grouping, and reflection strategies from  \ours. Table \ref{tab:abalationstudy} presents the ablated F1 results averaged from five-fold cross-validation. 
With gpt-3.5-turbo as the predictor LLM, the demonstration strategy is the most impactful. In contrast, both demonstration and grouping notably boost performance with mistral-7b, achieving the highest gains across four and five datasets (with~\emphOp{}). Overall, the performance gains are more pronounced for the `weaker' $\llmOp_p$ mistral-7b compared with gpt-3.5-turbo. Interestingly, \ours with mistral-7b ultimately outperforms with gpt-3.5-turbo by 1.5\%. This result also implies that open-sourced LLMs enhanced with insight distillation can be effective few-shot tabular classifiers, significantly reducing serving costs.
%
A closer analysis of the results with mistral-7b provides additional insights. For instance, removing grouping in the Bank, Blood, and Calhousing datasets leads to significant performance drop, indicating that summarized rules help manage class imbalance (statistics in Table~\ref{tab:datasets}) and better cope with numerical features (examples in Figure~\ref{fig:prompt-example-calhousing}). 

In summary, the three strategies consistently improve performance across nine datasets: demonstration aids in-context learning, grouping organizes data for rule summarization, and reflection refines results through iterative error correction. Together, these components ensure high accuracy and robustness. Appendix~\ref{appendix:comparative} presents results of alternative strategies explored when developing  \ours, further justifying the current model design.
\begin{figure}
    \centering
    \includegraphics[width=0.85\linewidth]{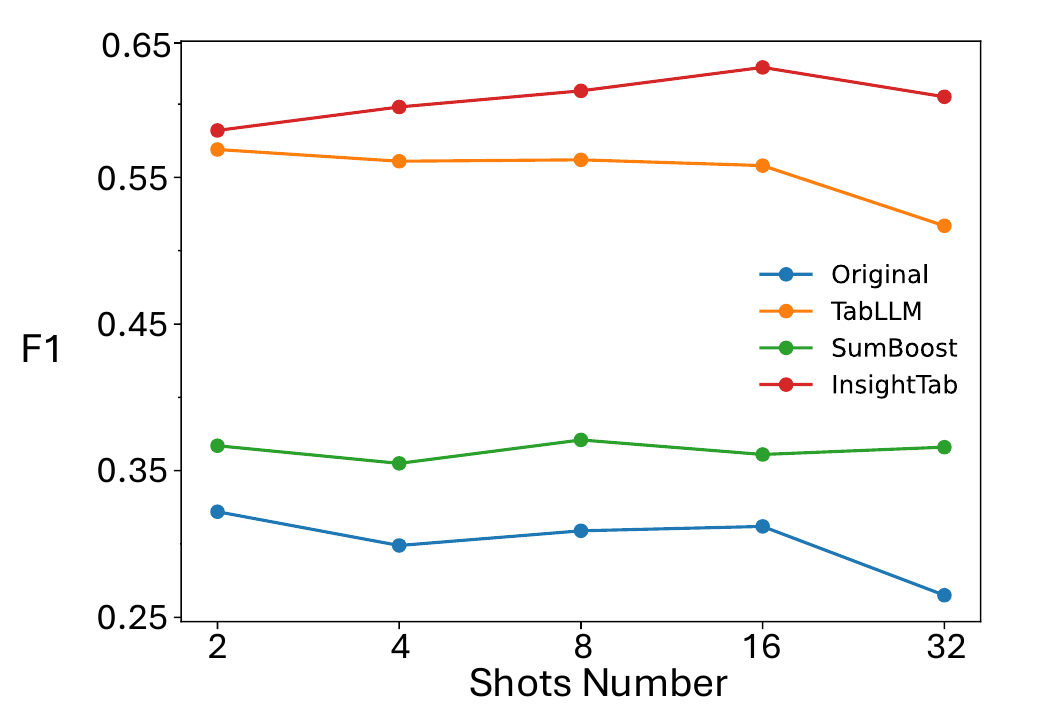}
    \caption{Mistral performance with 32 training samples across different number of shots (samples attached in the prompts). The results are averaged across 9 datasets. }
    \label{fig:shots-analysis}
\end{figure}

\begin{figure*}[t]
\begin{subfigure}{0.5\columnwidth}
\includegraphics[width=\textwidth]{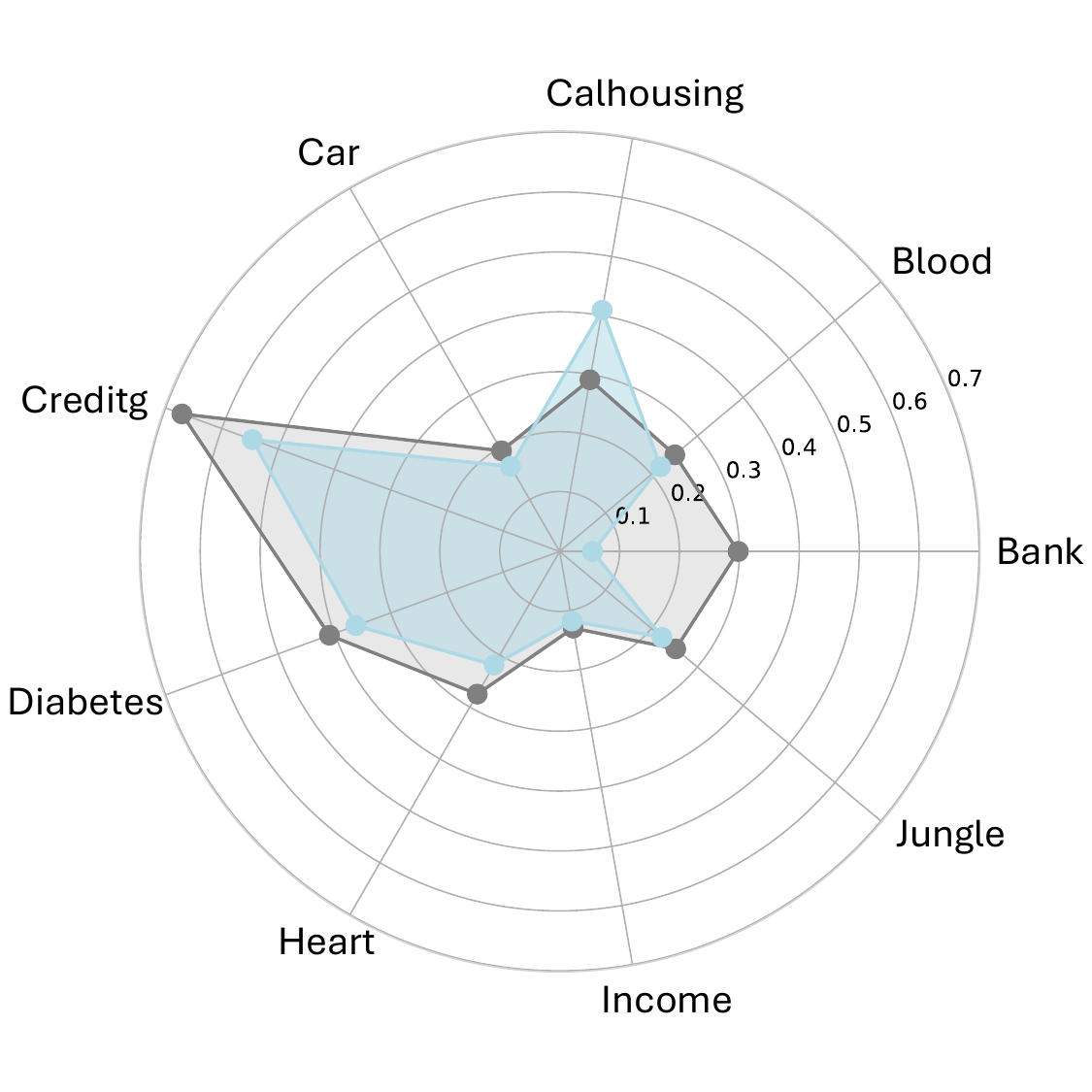}
\caption{Original}
\label{ori-mistral}
\end{subfigure}
\begin{subfigure}{0.5\columnwidth}
\includegraphics[width=\textwidth]{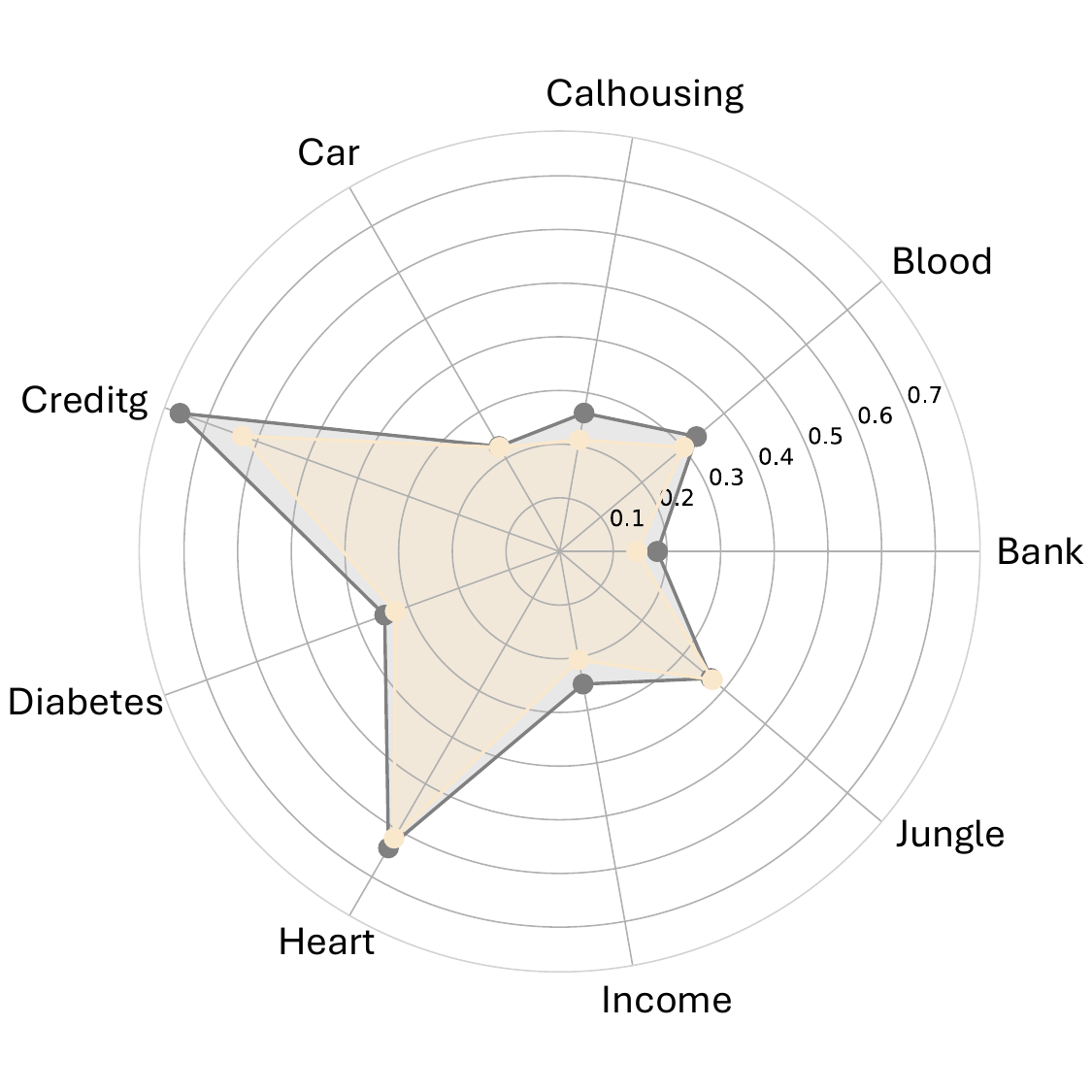}
\caption{\summaryboost}
\label{sum-mistral}
\end{subfigure}
\begin{subfigure}{0.5\columnwidth}
\includegraphics[width=\textwidth]{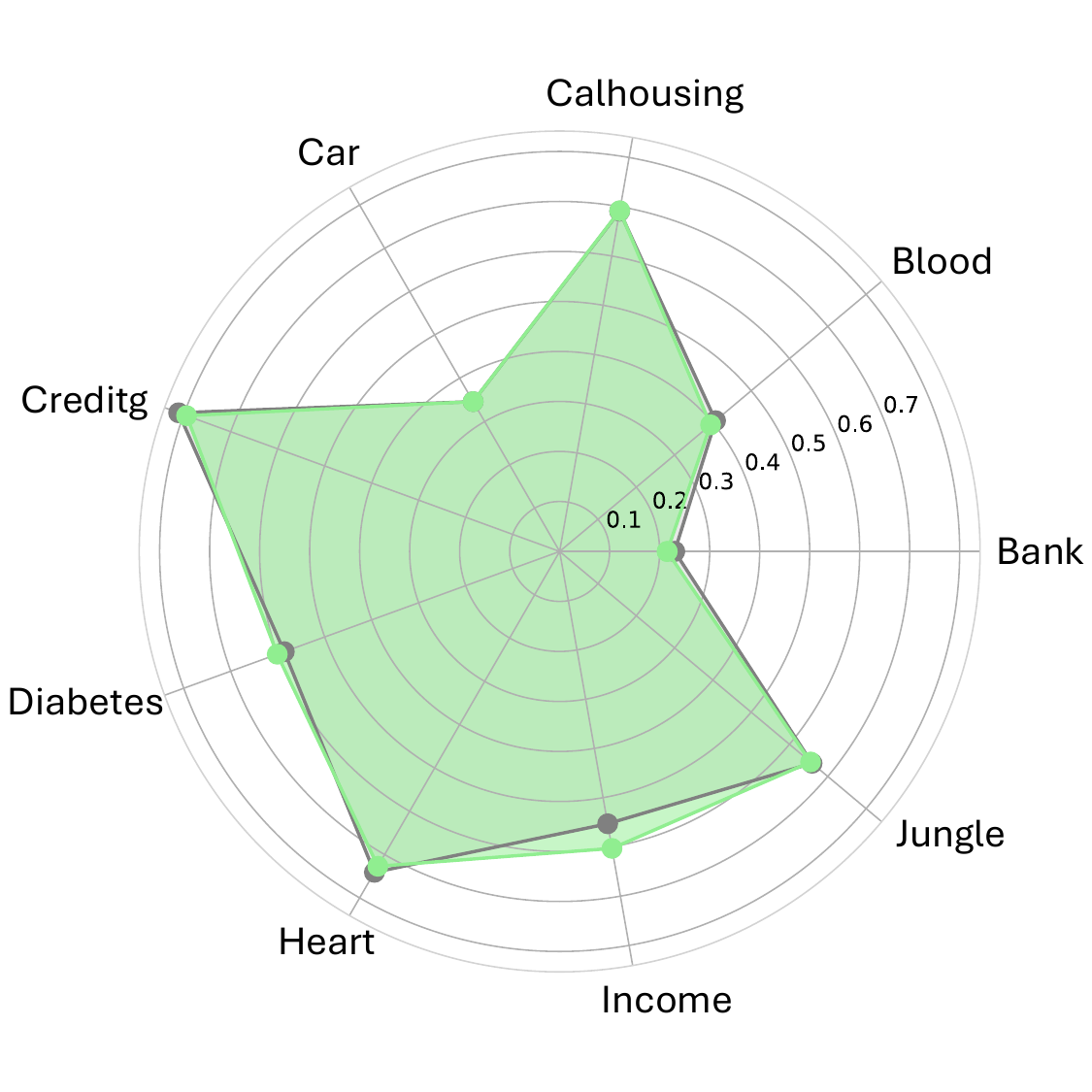}
\caption{TabLLM}
\label{tabllm-mistral}
\end{subfigure}
\begin{subfigure}{0.5\columnwidth}
\includegraphics[width=\textwidth]{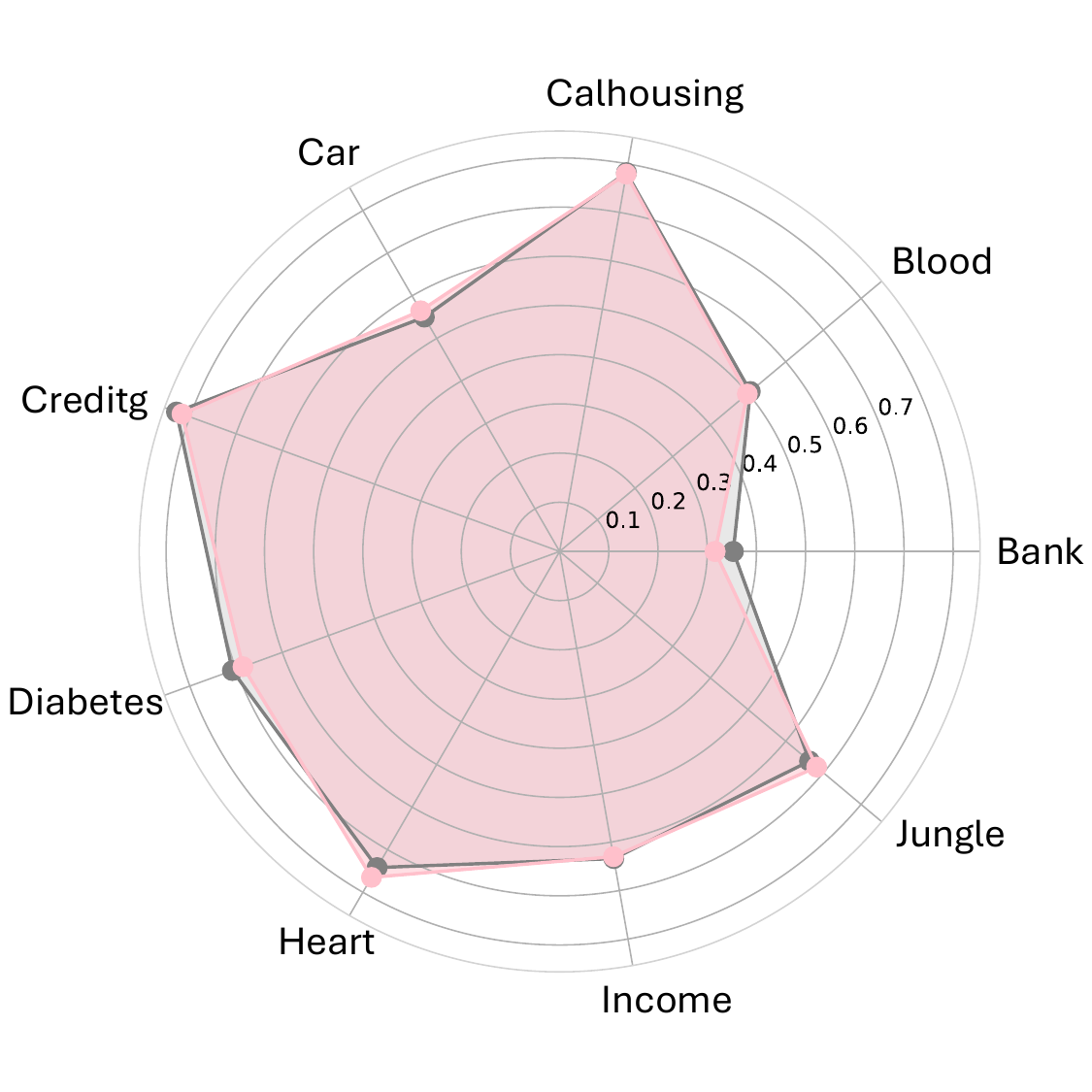}
\caption{\ours{}}
\label{ours-mistral}
\end{subfigure}
\begin{subfigure}{0.5\columnwidth}
\includegraphics[width=\textwidth]{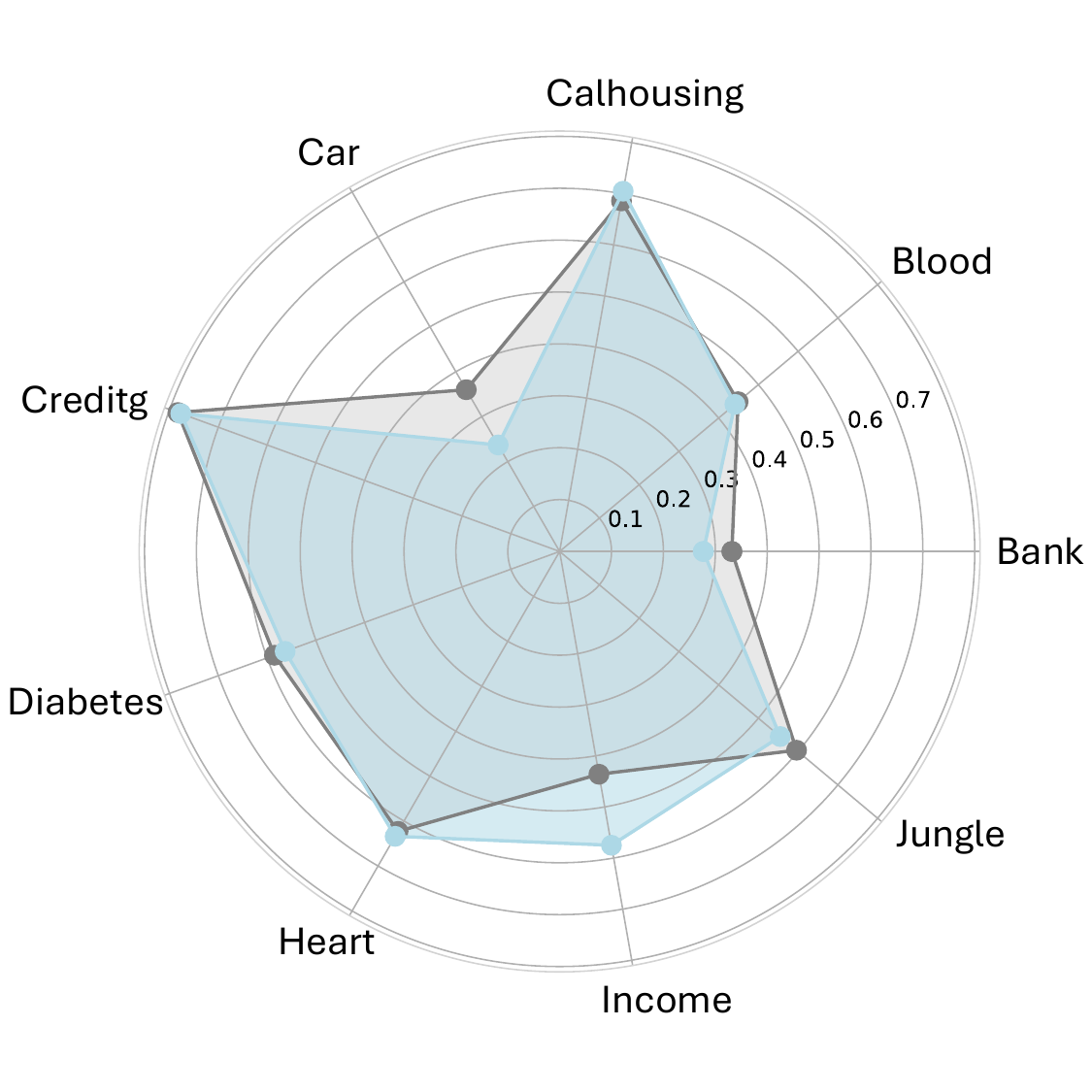}
\caption{Original}
\label{ori-gpt}
\end{subfigure}
\begin{subfigure}{0.5\columnwidth}
\includegraphics[width=\textwidth]{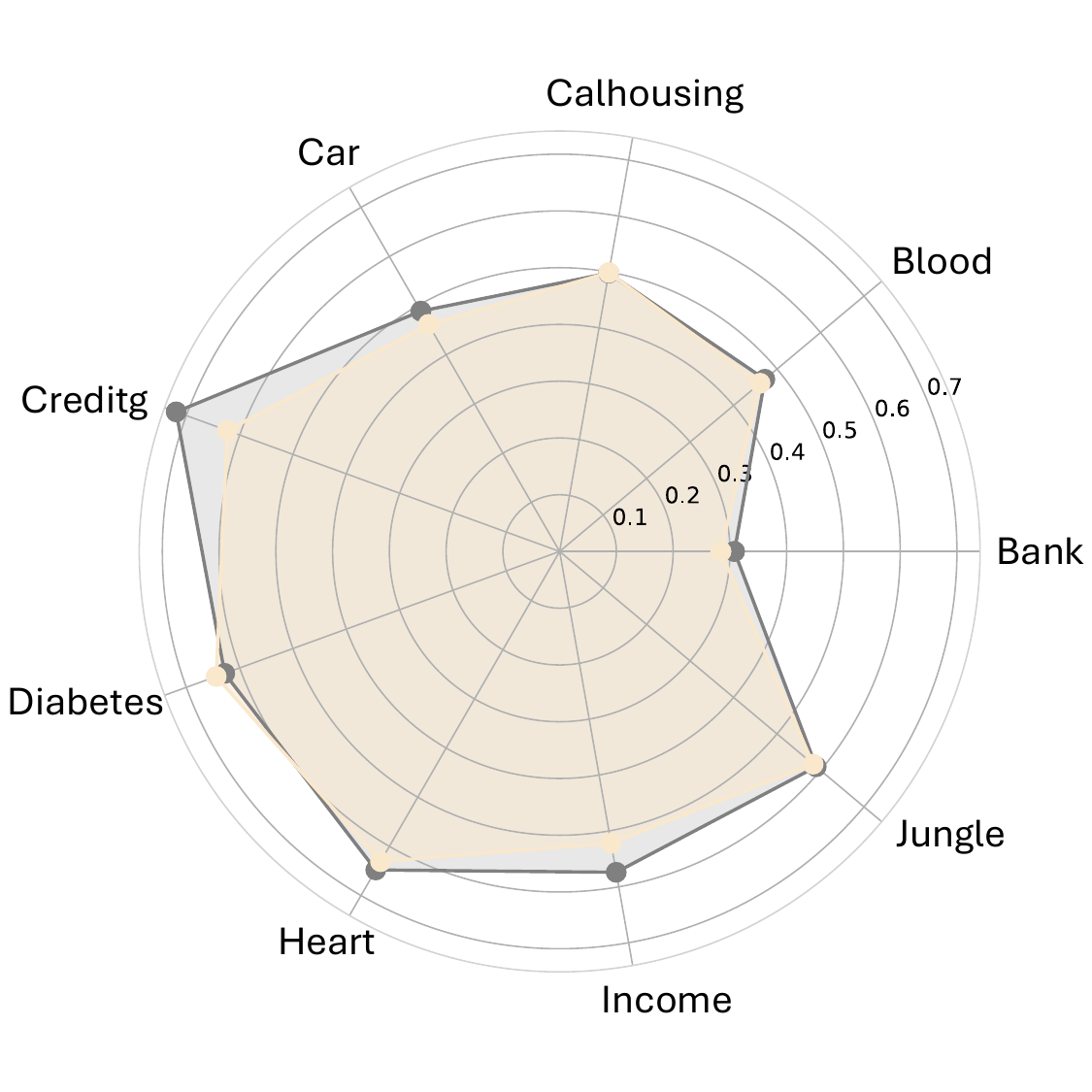}
\caption{\summaryboost}
\label{sum-gpt}
\end{subfigure}
\begin{subfigure}{0.5\columnwidth}
\includegraphics[width=\textwidth]{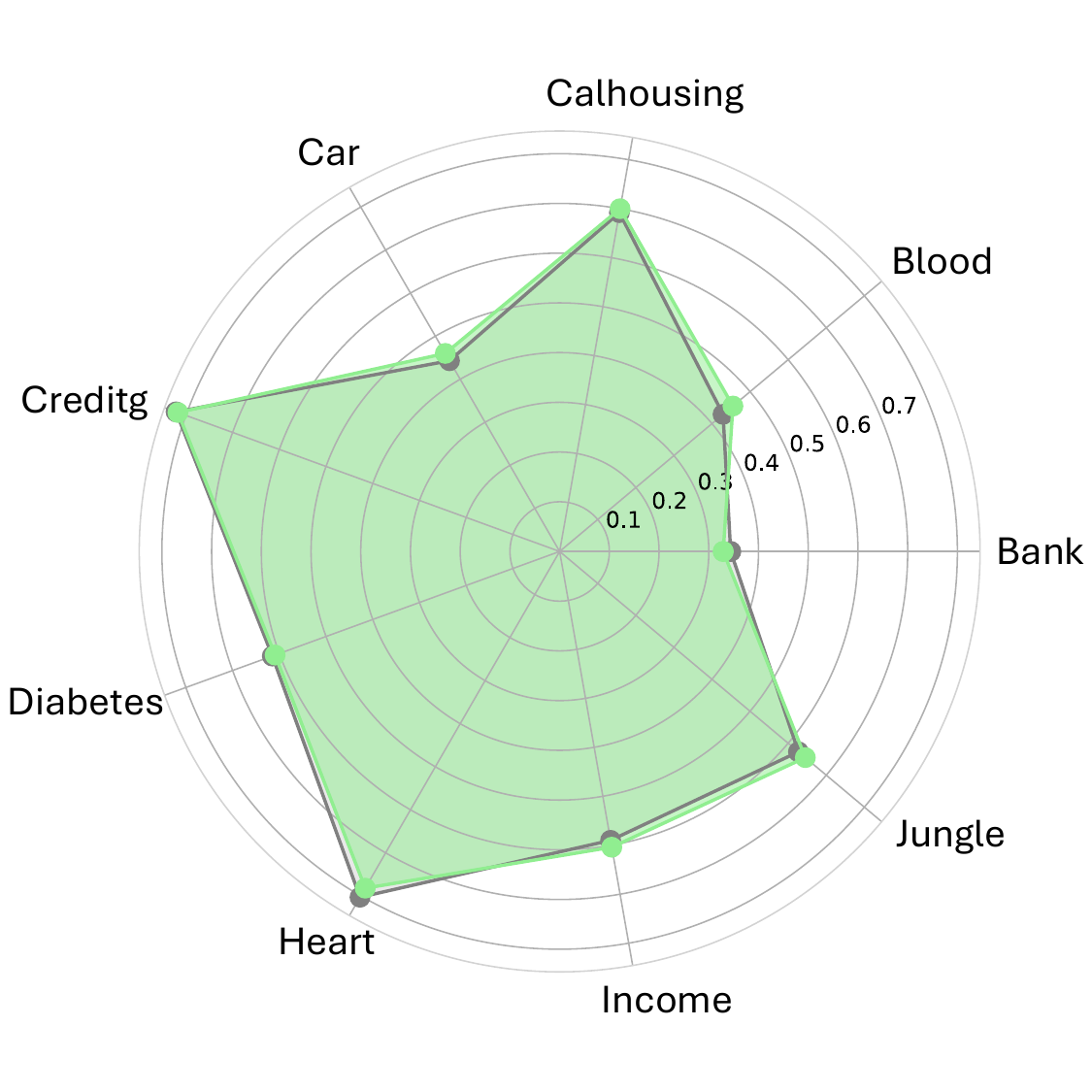}
\caption{TabLLM}
\label{tabllm-gpt}
\end{subfigure}
\begin{subfigure}{0.5\columnwidth}
\includegraphics[width=\textwidth]{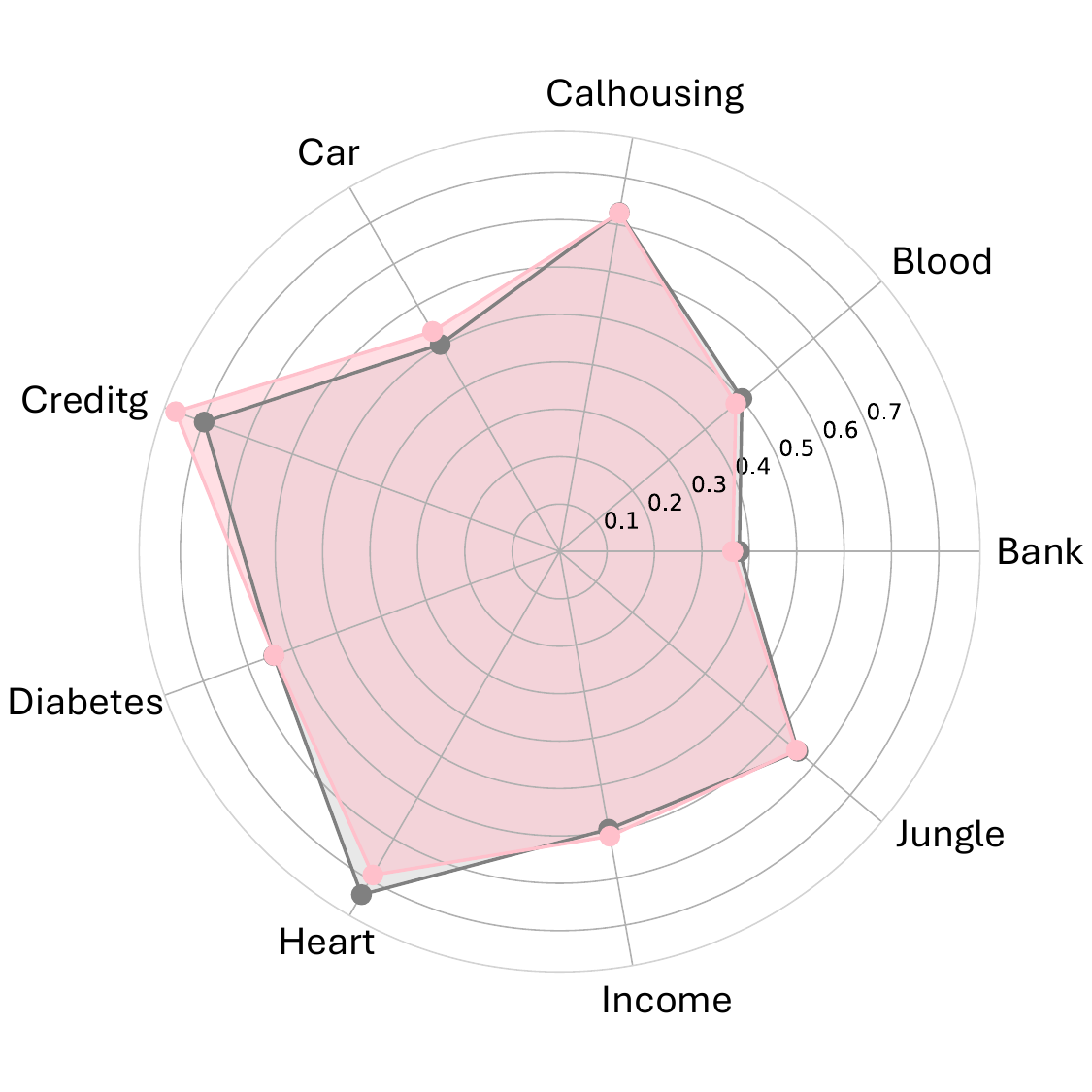}
\caption{\ours{}}
\label{ours-gpt}
\end{subfigure}
\caption{Position bias analysis (with 128 training samples and 16 shots) for four LLM-based methods using mistral-7b (upper) and gpt-3.5-turbo (lower). The grey/colored areas denote the F1 performance when the features of testing data are shuffled/non-shuffled, respectively.}
\label{position-bias-analysis}
\end{figure*}

\section{In-depth Analysis}
\label{extensive-aly}

We conduct a series of in-depth analysis on the task, which could deepen our understanding for this emerging paradigm. Additionally, we put detailed cost analysis in Appendix~\ref{appendix:cost}.

\subsection{Demonstration Shot Analysis}

We compare the performance of different models by adjusting the number of samples demonstrated in the prompts. Figure~\ref{fig:shots-analysis} shows the mistral performance (the results of gpt-3.5-turbo given in Appendix~\ref{appendix:gpt-3.5-shot}). We observe a consistent trend across all four models as the number of shots increases. The performance initially improves but then gradually declines after 16 shots. In other words, while adding more examples in the prompt can enhance performance up to a point, an excessive number of examples may lead to negative impacts. This might be due to that including too many examples causes the model to overfit to those specific instances, reducing its generalization to new, unseen data. However, methods incorporating rule generation (\eg \summaryboost) alleviate this issue compared to others (\eg TabLLM). Specifically, our model shows advantage in handling more in-context samples.
By integrating rule generation, our approach effectively mitigates the risk of overfitting, as it allows the model to extract and apply general principles rather than relying solely on individual examples, resulting in improved generalization and robustness in terms of large shot numbers.
\fboxsep=0.8pt 
\fboxrule=0pt 
\begin{figure*}[htp]
\centering
    \subfloat[SummaryBoost]{
    \begin{tcolorbox}[colback=white,colframe=black,width=0.35\textwidth, height=0.38\linewidth]
    \fontsize{8}{9}\selectfont
    1. Individuals who \colorbox{yellow}{work in} professional, managerial, or executive \colorbox{yellow}{positions} and have a \colorbox{yellow}{bachelor's degree or higher} are likely to earn \colorbox{green}{over \$50,000} annually.
    
    2. Individuals who \colorbox{yellow}{work full-time} (40 hours or more per week) in \colorbox{yellow}{occupations} such as engineering, healthcare, or finance, with \colorbox{yellow}{no capital gain or loss}, and are \colorbox{yellow}{married with a spouse} present are likely to earn \colorbox{green}{over \$50,000} annually.
    
    3. Individuals who \colorbox{yellow}{work in} office and administrative support roles for local government, have a \colorbox{yellow}{high school education}, \colorbox{yellow}{work 40 hours per week}, and are \colorbox{yellow}{divorced} are likely to earn \colorbox{green}{under \$50,000} annually.
    \end{tcolorbox}
    }
    \subfloat[InsightTab]{
    \begin{tcolorbox}[colback=white,colframe=black,width=0.6\textwidth, height=0.38\linewidth]
    \fontsize{8}{9}\selectfont
    1. \colorbox{yellow}{Higher education levels} (bachelor's degree or higher) generally correlate with earning \colorbox{green}{more than \$50,000 per year}. 
    
    2. \colorbox{yellow}{Occupations} in professional specialties or management, or \colorbox{yellow}{ownership} of an incorporated business, are likely to earn \colorbox{green}{above \$50,000}. 
    
    3. \colorbox{yellow}{Working full-time hours} (40 hours per week or more) is commonly associated with \colorbox{green}{higher earnings}. 
    
    4. \colorbox{yellow}{Significant capital gains} in a year are a strong indicator of earning \colorbox{green}{more than \$50,000}. 
    
    5. \colorbox{yellow}{Marital status} as married and relation to head of household as husband frequently appear in profiles of those earning \colorbox{green}{more than \$50,000}. 
    
    6. \colorbox{yellow}{Working in the private sector}, especially in technology, support, or professional fields, often correlates with incomes \colorbox{green}{above \$50,000}. 
    
    7. \colorbox{yellow}{Lower educational attainment and employment} in sectors like agriculture, forestry, fisheries, or non-specialized government roles tend to correlate with incomes \colorbox{green}{below \$50,000}.
    \end{tcolorbox}
    }
    \caption{Rules summarized by SummaryBoost (baseline) and InsightTab (ours).}
    \label{fig:rules-summaryboost-insighttab}
\end{figure*}
\begin{figure}[t]
\begin{subfigure}{0.85\columnwidth}
\includegraphics[width=0.95\textwidth,height=40mm]{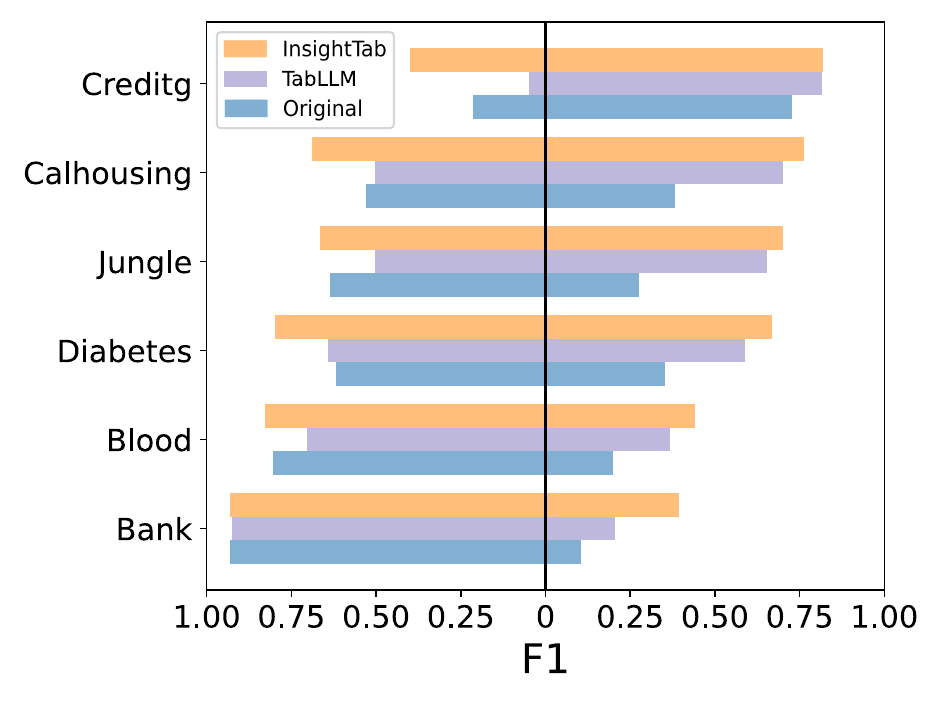}
\caption{mistral-7b}
\label{mistral-class-analysis}
\end{subfigure}
\begin{subfigure}{0.85\columnwidth}
\includegraphics[width=0.95\textwidth,height=40mm]{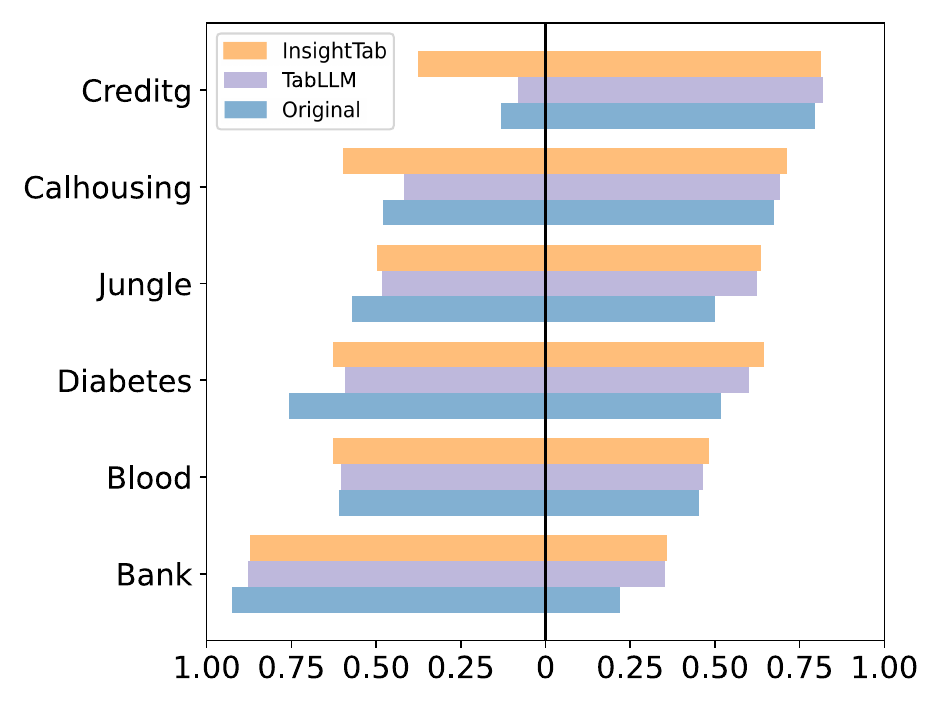}
\caption{gpt-3.5-turbo}
\label{gpt-class-analysis}
\end{subfigure}
\caption{F1 scores for True and False classes across datasets (\#32 training samples, 16 shots): False class on the left, True class on the right. }
\label{category-alys}
\vspace{-4mm}
\end{figure}
\subsection{Bias Analysis}

We evaluate our model's performance on position and class biases against baseline models. 

\stitle{Position bias analysis}
We first investigate the model's performance in addressing position bias. Specifically, we randomly shuffle the order of columns of every testing sample to assess whether the models can handle this discrepancy between the training and testing sets. We report the performance in Figure \ref{position-bias-analysis}. We observe that both serialization (TabLLM vs. Original) and rule summarization (\summaryboost vs. Original) improve the model's ability to handle randomly shuffled testing cases, as shuffling has influenced the performance on many datasets (\eg Income) greatly in the Original method as shown in Figs.~\ref{ori-mistral} \& \ref{ori-gpt}. Since the Original method may rely solely on memorizing specific feature order rules, lacking a meaningful semantic structure. In comparison, \ours{} demonstrates better robustness on feature order changes. The average performance change due to the shifting of the test data is significantly lower in our model compared to the \summaryboost and Original baselines. After the shift, the performance of GPT-3.5-Turbo in our model reaches 64.2\%, compared to 36.9\% with \summaryboost and 31.4\% with the Original. Additionally, compared with TabLLM, \ours{} demonstrates better performance across different datasets. This demonstrates that our model effectively learns the semantic relationships between features via multifaceted serialization, rather than simply memorizing feature positions. 

\stitle{Unbalanced class analysis}
To assess \ours's performance on unbalanced datasets, Figure~\ref{category-alys} compares F1 scores for True and False classes across datasets with increasing True/False ratios. We find the predictions of Original are heavily biased to class distributions. For example, the   F1 score of True label on Bank with mistral-7b (see Figure \ref{mistral-class-analysis}) is extremely low as 0.11 compared with 0.93 for the False label, because the True class constitutes only 11.7\% of the dataset. However, compared with Original and TabLLM, this situation alleviates in \ours{}, showcasing more balanced results for both classes. In the Creditg dataset, the True/False F1 score of \ours{} is 0.82/0.41 with mistral-7b, compared with 0.81/0.06 by TabLLM. This confirms our model's robustness on unbalanced datasets by ensuring representation of both classes via sampling.

\subsection{Case Study}
\par We showcase rules summarized by SummaryBoost and InsightTab to demonstrate the differences between the two methods in terms of data insights (see Figure \ref{fig:rules-summaryboost-insighttab}). We select rules summarized in the Income dataset considering its scale and quality. We obtain several findings via the case study.

\par\noindent\textbf{The rules offer a high-level interpretation of the feature-label relationship.}  We use yellow to highlight features and green to highlight labels, emphasizing the feature-label mapping in the summarized rules. A shared characteristic of rules generated by SummaryBoost and InsightTab is their essence as compressed representations of feature-label mappings. These rules precisely and clearly describe the relationships between specific features and labels, enabling LLMs (and humans) to efficiently interpret and perform tabular predictions.

\par\noindent\textbf{Rules summarized by InsightTab are clearer and more understandable.} A key characteristic of InsightTab's rules is their one-to-one mapping, each focusing on a specific feature, while SummaryBoost generates many-to-one mappings, with each capturing multiple features for an individual. One-to-one mappings are simpler and more interpretable for weak table predictors, and the independence of the generated rules allows predictors to handle unseen feature combinations. In such cases, the predictor LLM is better equipped to prioritize important features and accurately predict labels based on the summarized rules.

    
\section{Conclusion}
We propose \ours{}, a novel method for accuarte and robust LLM-based few-shot tabular classification. By distilling training data into actionable insights, \ours{} adapts LLMs to tabular tasks via the \textit{divide-and-conquer}, \textit{easy-first}, and \textit{reflective learning} principles, implemented through \verb|group|, \verb|rank|, and \verb|summarize| operators. They enable grouping samples to derive rules, selecting representative instances, and learning from errors to enhance performance. 
We conduct experiments on nine datasets, comparing our method with state-of-the-art approaches. Our method consistently achieves superior average performance across all nine datasets. Extensive analysis shows that our model is more robust to positional and class biases.

\section*{Limitations}
While \ours{} excels in few-shot tabular classification, its two-stage inference process incurs some cost. However, by leveraging off-the-shelf LLMs and few-shot demonstrations, it significantly reduces training time and expense compared to non-few-shot methods. It relies on a powerful language model \( \llmOp_s \) to summarize rules from tabular data, but this cost is minimal when amortized over many predictions and can be further reduced by reusing rules. Additionally, applying these rules to a more efficient open-source model \( \llmOp_p \) lowers serving costs without compromising performance.


Another limitation is our focus on the few-shot setting. When a sufficiently large training dataset is available, traditional methods like \xgb are still more effective. Performance comparisons under full training data are discussed in Appendix~\ref{app:extra-exp-fulltrain}. 
Finally, evaluating LLM-based approaches faces challenges related to data contamination, which has been verified on tabular data~\cite{bordt2024elephants}. This can lead to overestimated performance due to memorization of target values. We have made significant efforts to minimize this impact, such as employing five-fold cross-validation and studying feature position bias. Specifically, our position bias analysis suggests that both serialization and rule summarization, which can be viewed as perturbations to raw data~\cite{li2024perteval}, enhance robustness of model behaviors. However, future work should follow this direction and explore whether our model remains effective when column names lack semantic meaning.


\section*{Ethics Statement}
We utilize the full training and testing data from public databases (details provided in Appendix \ref{appendix:data}) and ensure that no personal data is included in the datasets.





\bibliography{references}
\begin{appendix}

\section{Mined Rules}
\label{mined-rules}
\begin{table}[tbh]
  \small
  \begin{tabularx}{.47\textwidth}{X}
    \toprule
        1. Higher education levels (bachelor's degree or higher) generally correlate with earning more than \$50K per year. 
        
        2. Significant capital gains in a year are a strong indicator of earning more than \$50K. 
        
        3. Lower educational attainment and employment in sectors like agriculture, forestry, fisheries, or non-specialized government roles tend to correlate with incomes below \$50K. \\
    \bottomrule
  \end{tabularx}
\caption{Illustration of mined rules for Income data.}
\label{tab:rules}
\end{table}

Table~\ref{tab:rules} illustrates three mined rules related to the features presented in Fig.~\ref{fig:model-structure}  for income prediction. 
As could be seen, these rules consist of both general knowledge from LLMs and task-specific details distilled from data, \eg bachelor's degree or higher, capital gains, and employment. The combination of the two worlds enables the prediction LLM leveraging its capacity in a task-specific context for tabular prediction. 
The complete list of mined rules, multifaceted serialization prompts, and the prompts for rule summarization and merge could be found in Appendix~\ref{appendix:prompts}.

\section{Experimental Setup Details}
\label{appendix:exp}

\subsection{Datasets}
\label{appendix:data}
We adopt nine tabular datasets with varying sizes, diverse features, and different classification tasks for evaluation, first proposed by~\citet{Hegselmann2022TabLLMFC}. 
Specifically, these datasets include \textbf{Bank} (predict whether a customer subscribed to a term deposit or not), \textbf{Blood} (predict whether a person will return for another blood donation), \textbf{California Housing / Calhousing} (predict the median house value in each district), \textbf{Car} (evaluate the safety state level of a car), \textbf{Creditg} (predict whether a person is at high or low probability for credit card approval), \textbf{Diabetes} (predict whether a person has diabetes), \textbf{Heart} (predict whether a person has coronary artery disease), \textbf{Income} (predict whether each person has an annual income over \$50,000) and \textbf{Jungle} (predict whether the white player will win in a jungle chess game). The statistics of each dataset are summarized in Table~\ref{tab:datasets} and the raw data is available at \url{https://github.com/clinicalml/TabLLM}. 

\begin{table}[]
    \small
    \centering
    \begin{tabular}{cccccc}
    \toprule
        Dataset & \#Rows & \#F & Classes & Cls Dist. (\%) \\
    \midrule
       Bank   & 45,211 & 16 & binary & 11.7/88.3\\
       Blood  & 748 & 4 & binary & 23.8/76.2 \\
       Calhousing & 20,640 &  8 & binary & 50.0/50.0\\
       Car & 1,728 &  8 & 0/1/2/3   & 70/22/4/4  \\
       Creditg & 1,000 & 20 & binary & 70.0/30.0\\
       Diabetes & 768 & 8 & binary & 34.9/65.1 \\
       Heart & 918 & 11 & binary & 55.3/44.7\\
       Income & 48,842 & 14 & binary & 23.9/76.1\\
       Jungle &44,819 &6 & binary & 48.5/51.5\\
    \bottomrule
    \end{tabular}
        \caption{Dataset statistics, where `\#F' and `Cls Dist.' denote the number of features and Class Distribution, respectively.}
    \label{tab:datasets}
\end{table}

\subsection{Baselines}
\label{appendix:baselines}
We compare our method \ours with a broad spectrum of methods, with a focus on few-shot approaches:
\begin{itemize}
    \item \textbf{\xgb}~\cite{Chen2016XGBoostAS} is a classic tree-ensemble model that remains one of the most effective methods for tabular prediction in general~\cite{Grinsztajn2022WhyDT}. However, it inherently relies on a substantial amount of data to identify optimal splitting features and values, which typically leads to its ineffectiveness in few-shot settings.
    \item \textbf{\tabpfn}~\cite{Hollmann2022TabPFNAT} is a trained Transformer that performs in-context learning to do supervised classification for small tabular datasets  without requiring further parameter updates.
    \item \textbf{Original} is the most straightforward LLM baseline, which classifies tabular data by feeding the features as a textual prompt into LLMs. Each tabular data instance is serialized as \verb|{column_name}:{value}| separated by a space between columns.
    \item \textbf{Tablet}~\cite{Slack2023TABLETLF} employs automatic generation of dataset-specific natural language instructions to boost the tabular classification performance with LLMs. We use prototypes instructions as they are reported preforming better in the original paper.
    \item \textbf{TabLLM}~\cite{Hegselmann2022TabLLMFC} is an LLM-based tabular classification model designed for few-shot settings. It explores extensively on the different serialization methods and finds that the \emph{Text Template} approach performs well in general.
    \item \textbf{\summaryboost}~\cite{Manikandan2023LanguageMA} is the most state-of-the-art LLM method for tabular classification. It leverages a boosting strategy to sample and summarize training data for LLM  instruction optimization.
\end{itemize}

\begin{table*}[!h]
\footnotesize
\centering
\begin{tabular}{c c c c c c c c c c}
\toprule
    Method & Bank & Blood & Calhou. & Car & Creditg & Diabe. & Heart & Income & Jungle  \\
\midrule
    \multicolumn{10}{c}{$n=16$} \\
    \xgb & 0.0 & 27.3 & 66.4 & 33.2 & 80.1 & 31.2 & 69.3 & 10.5 & 52.8\\
    \tabpfn & 1.3 & 21.6 & 72.1 & 22.1 & \textbf{82.2} & 29.2 & 79.0 & 0.4 & 41.6\\
    \baseline & 5.6/19.6 & 29.0/45.3 & 44.7/67.0 & 20.1/22.9 & 72.9/73.8 & 35.3/43.1 & 27.4/55.5 & 13.2/44.9 & 12.4/43.8\\
    \tablet & 11.3/31.8 & 27.6/43.8 & 53.4/62.1 & 26.9/36.2 & 66.1/71.2 & 40.5/61.9 & 48.4/77.1 & 23.1/54.7 & 32.3/62.8\\
    \tabllm & 21.1/32.9 & 38.7/44.0 & 68.7/69.4 & 35.2/\underline{50.0} & 80.2/\textbf{82.2} & 60.0/62.1 & 73.7/\underline{79.9} & 58.5/59.6 & 57.8/61.5\\
    \summaryboost & 16.2/28.7 & 30.1/35.6 & 27.9/57.1 & 23.4/41.2 & 64.3/49.9 & 36.3/59.3 & 60.4/62.6 & 26.0/52.6 & 39.1/60.0\\
    \ours & \textbf{48.8}/\underline{41.7} & \textbf{62.9}/\underline{54.9} & \textbf{79.7}/\underline{75.8} & 48.4/\textbf{51.5} & 80.4/80.9 & \textbf{76.4}/\underline{68.2} & \textbf{80.0}/76.7 & \textbf{67.5}/\underline{61.0} & \textbf{69.4}/\underline{64.9}\\
\midrule
    \multicolumn{10}{c}{$n=32$} \\
    \xgb & 14.9 & 23.5 & 66.5 & 33.6 & 76.5 & 52.2 & 74.6 & 39.8 & 69.6\\
    \tabpfn & 16.6 & 9.5 & \underline{76.0} & 23.8 & \textbf{81.8} & 37.2 & \textbf{79.8} & 8.6 & \underline{70.1}\\
    \baseline & 10.4/31.5 & 19.9/45.5 & 38.3/67.1 & 19.7/35.1 & 72.7/80.2 & 35.3/57.9 & 34.9/60.7 & 13.9/44.7 & 27.5/58.1\\
    \tablet & 13.6/34.0 & 33.4/42.6 & 54.9/65.3 & 25.9/32.3 & 66.6/77.6 & 39.3/61.7 & 43.8/77.3 & 25.4/55.3 & 42.6/63.0\\
    \tabllm & 20.6/35.3 & 36.6/\underline{46.4} & 70.1/69.1 & 36.8/\underline{45.5} & 81.6/\textbf{81.8} & 58.9/60.1 & 73.8/78.6 & 58.2/61.6 & 65.3/62.4\\
    \summaryboost & 19.2/30.2 & 35.4/46.3 & 20.4/50.1 & 22.9/44.6 & 60.9/70.8 & 38.1/61.0 & 64.7/62.9 & 26.7/\underline{62.1} & 36.8/58.8\\
    \ours & \textbf{39.5}/\underline{35.8} & 44.1/\textbf{48.3} & \textbf{76.4}/71.2 & 43.9/\textbf{49.1} & \textbf{81.8}/81.4 & \textbf{66.8}/\underline{64.6} & 77.5/\underline{78.8} & 62.0/\textbf{64.7} & \textbf{70.2}/63.6\\
\midrule
    \multicolumn{10}{c}{$n=64$} \\
    \xgb & 32.3 & 39.2 & 72.6 & 43.3 & 77.3 & 58.5 & \underline{81.9} & 36.2 & \underline{72.2}\\
    \tabpfn & 21.4 & 11.6 & \textbf{80.0} & 28.5 & 80.4 & 57.7 & \textbf{83.7} & 35.0 & \textbf{72.3}\\
    \baseline & 22.1/31.4 & 21.5/45.7 & 36.7/70.2 & 21.5/37.4 & 70.0/80.4 & 35.1/56.9 & 32.4/56.1 & 15.3/46.2 & 30.0/57.0\\
    \tablet & 5.9/27.4 & 28.0/42.0 & 55.3/63.8 & 24.3/34.4 & 66.6/76.0 & 37.0/61.4 & 37.4/74.8 & 22.4/56.4 & 40.7/62.1\\
    \tabllm & 22.1/35.6 & 36.1/45.9 & 68.9/68.8 & 34.6/45.2 & 80.7/81.8 & 59.1/61.5 & 74.7/79.3 & 58.9/59.3 & 66.2/63.5\\
    \summaryboost & 14.8/29.8 & 32.3/45.2 & 18.2/43.3 & 22.4/\underline{47.8} & 56.0/66.0 & 42.8/63.1 & 61.3/64.6 & 21.7/60.9 & 36.8/59.0\\
    \ours & \textbf{40.8}/\underline{39.7} & \textbf{49.5}/\underline{49.3} & \underline{74.6}/69.7 & 47.2/\textbf{52.2} & \underline{82.6}/\textbf{83.5} & \textbf{64.4}/\underline{63.3} & 75.0/81.2 & \textbf{72.9}/\underline{62.0} & 71.6/63.6\\
\midrule
    \multicolumn{10}{c}{$n=128$} \\
    \xgb & \underline{36.2} & 41.4 & 76.9 & \underline{50.7} & 79.7 & 58.3 & \textbf{85.9} & 49.2 & \underline{76.8}\\
    \tabpfn & 18.0 & 27.8 & \textbf{83.2} & 47.7 & 80.6 & 62.5 & \underline{84.8} & 36.8 & \textbf{77.2}\\
    \baseline & 29.8/33.2 & 25.1/44.9 & 29.1/68.6 & 19.4/36.0 & 68.1/78.3 & 40.9/58.5 & 27.5/62.3 & 13.0/43.6 & 25.3/59.6\\
    \tablet & 12.2/31.5 & 30.0/46.5 & 55.9/61.0 & 24.5/35.1 & 67.5/71.8 & 42.4/59.3 & 39.4/77.3 & 25.2/55.6 & 40.3/63.1\\
    \tabllm & 23.0/34.4 & 40.7/42.9 & 69.1/69.2 & 34.6/44.3 & 81.1/\underline{82.1} & 58.6/61.5 & 74.1/80.3 & 58.3/59.0 & 65.9/62.7\\
    \summaryboost & 18.2/30.8 & 33.3/47.2 & 26.2/49.8 & 22.6/48.9 & 75.3/71.9 & 34.7/62.8 & 63.8/64.8 & 25.1/57.4 & 36.8/59.0\\
    \ours & 35.3/\textbf{37.9} & \textbf{50.6}/\underline{50.2} & \underline{78.2}/72.6 & \textbf{55.0}/50.4 & \textbf{82.9}/79.8 & \textbf{70.8}/\underline{64.2} & 74.2/83.6 & \textbf{63.4}/\underline{59.5} & 66.3/65.6\\
\midrule 
    \multicolumn{10}{c}{$n=\mbox{all}$} \\
    \xgb & \textbf{54.5} & 42.4 & \textbf{90.5} & \textbf{97.7} & \underline{82.8} & 60.9 & \textbf{87.6} & \textbf{72.4} & \textbf{89.1}\\
    \tabpfn &  \multicolumn{9}{c}{/}\\
    \baseline & 17.8/31.2 & 24.1/44.8 & 47.7/68.8 & 21.3/34.4 & 71.4/78.9 & 40.4/57.0 & 26.9/58.0 & 12.0/47.6 & 26.9/55.2\\
    \tablet & 10.1/33.7 & 29.3/46.0 & 57.9/66.1 & 24.2/33.5 & 60.2/70.5 & 39.1/61.5 & 36.8/77.1 & 28.5/55.8 & 25.3/63.4\\
    \tabllm & 24.1/34.4 & 40.1/45.0 & 69.0/69.1 & 33.1/43.4 & 80.1/82.2 & 57.9/61.5 & 73.5/78.9 & 55.9/59.2 & 65.7/62.3\\
    \summaryboost & 20.5/32.0 & 34.2/45.3 & 48.9/67.7 & 26.7/46.1 & 79.9/78.7 & 41.0/58.4 & 65.4/79.1 & 30.2/52.9 & 28.6/61.1\\
    \ours & 36.8/\underline{36.9} & \underline{47.6}/\textbf{48.8} & \underline{71.6}/69.4 & 47.6/\underline{52.9} & \textbf{83.7}/79.8 & \textbf{65.5}/\underline{62.7} & 75.6/\underline{80.3} & \underline{64.2}/59.5 & \underline{70.8}/63.8\\
\bottomrule
    \end{tabular}   
 \caption{Overall F1 performance on the public tabular datasets with different number of training samples $n$. For each LLM-based model, we report the performance based on mistral-7b/gpt-3.5-turbo using 16 shots. The best and second-best results are \textbf{bolded} and \underline{underlined}. }
    \label{tab:main-detailed}
\end{table*}

\begin{table*}[htp]
\centering
\small
\begin{tabular}{cc ccccccccc c}
\toprule
$\llmOp_p$ & Variant & Bank & Blood & Calhou. & Car & Creditg & Diabe. & Heart & Income & Jungle & Avg \\
\midrule
\multirow{5}{*}{mistral-7b}  
& R demos.  & 10.4 & 12.4 & 17.3 & 39.2 & 35.1 & 42.2 & 56.0 & \emphOp{32.4} & \emphOp{13.3} & 28.7 \\
& SV demos. & \emphOp{10.2} & 13.2 & 16.8 & \emphOp{37.7} & \underline{35.8} & \emphOp{41.5} & \emphOp{53.3} & 33.8 & 14.7 & \emphOp{28.6} \\
& LR rule  & 11.0 & \emphOp{11.1} & \underline{17.4} & \underline{39.6} & 35.5 & 43.9 & 54.3 & 35.6 & 15.8 & 29.4 \\
& Cocktail  & \underline{12.0} & \underline{15.9} & \emphOp{14.1} & 38.4 & \emphOp{34.8} & \underline{54.3} & \underline{66.7} & \underline{50.3} & \underline{18.2} & \underline{33.9} \\
& InsightTab & \textbf{35.3} & \textbf{50.6} & \textbf{78.2} & \textbf{55.0} & \textbf{82.9} & \textbf{70.8} & \textbf{74.2} & \textbf{63.4} & \textbf{66.3} & \textbf{64.1} \\ \midrule 
\multirow{5}{*}{\makecell{gpt-3.5 \\ -turbo}} 
& R demos.  & 26.6 & 42.8 & \textbf{73.9} & 41.6 & \underline{68.3} & 62.2 & \emphOp{68.3} & 52.2 & \textbf{65.8} & \underline{55.7} \\
& SV demos. & \underline{27.4} & \emphOp{40.5} & 71.4 & 40.1 & 67.8 & 61.2 & 69.9 & 54.5 & 64.4 & 55.2 \\
& LR rule  & \emphOp{25.4} & 41.4 & 71.1 & \underline{42.3} & 68.2 & \underline{62.5} & 68.9 & \emphOp{37.0} & 62.3 & \emphOp{53.2} \\
& Cocktail  & 26.1 & \underline{52.6} & \emphOp{70.5} & \emphOp{39.7} & \emphOp{47.5} & \emphOp{59.6} & \underline{71.9} & \underline{59.1} & \emphOp{59.2} & 54.0 \\
& InsightTab & \textbf{37.9} & \textbf{50.2} & \underline{72.6} & \textbf{50.4} & \textbf{79.8} & \textbf{64.2} & \textbf{83.}6 & \textbf{59.5} & \underline{65.6} & \textbf{62.6} \\
\bottomrule
\end{tabular}
\caption{Results of alternative strategies with 128 training samples, 16 shots, and averaged F1 scores (\%). The best, second-best and worst results within each group are in \textbf{bold}, \underline{underlined}, and with \emphOp{}, respectively.}
\label{tab:ablation-appendix}
\end{table*}

\begin{table*}[htp]
\centering
\footnotesize

\begin{tabular}{lcccccccccc}
\toprule
\textbf{}                            & \textbf{Bank} & \textbf{Blood} & \textbf{Calhous.} & \textbf{Car} & \textbf{Creditg} & \textbf{Diabetes} & \textbf{Heart} & \textbf{Income} & \textbf{Jungle}        \\ \midrule
Rule Generation (gpt-4o-mini)        & 0.02          & 0.03          & 0.01             & 0.02        & 0.01            & 0.01              & 0.02          & 0.02           & 0.01         \\
Rule Generation (gpt-4-turbo)        & 1.58          & 1.66          & 1.87             & 1.97        & 1.80            & 1.67              & 1.70          & 1.80           & 1.72          \\
Prediction (gpt-3.5-turbo)           & 0.69          & 0.28          & 0.42             & 0.26        & 0.29            & 0.22              & 0.30          & 0.73           & 0.84        \\
\summaryboost (gpt-4-turbo)           & 12.3          & 14.5          & 15.8             & 14.1        & 11.4            & 10.4              & 15.7          & 18.9           & 17.5           \\
\bottomrule
\end{tabular}
\caption{Overall cost (USD) of InsightTab. Each number is reported using 5-fold validation on 128 training examples, 16 number of shots.}
\label{tab:cost}
\end{table*}

\begin{table*}[htp]
\centering
\footnotesize

\begin{tabular}{lc}
\toprule
\textbf{}                            & Time \\
\midrule
Rule Generation (gpt-4o-mini) & ~2min 20s   \\
Rule Generation (gpt-4-turbo) & ~2min 20s   \\
Prediction (gpt-3.5-turbo) & ~5min per 10k testing samples\\
\summaryboost (gpt-4-turbo)  & ~20min per 10k testing samples \\ 
\bottomrule
\end{tabular}
\caption{Time cost of \ours and baselines.}
\label{tab:time_cost}
\end{table*}

\subsection{Implementation Details}
\label{appendix:parameters}
We grid-search hyper-parameters for \xgb (\ie n\_estimators in \{10, 50, 100, 200\}, max\_depth in \{3, 4, 5, 6\}, learning\_rate in \{0.01, 0.03, 0.1, 0.3\}) and \tabpfn (\ie n\_ensemble in \{2, 4, 8, 16, 32, 64\}). We adopt the other parameters the same from the default settings in the XGBoost implementations in sklearn. We set the default value for the number of boosting rounds to 100, which is the same as that in \emph{xgboost.XGBRFClassifier}. The rest of parameters are listed as follows:

\begin{lstlisting}[language=Python]
XGBClassifier(
    base_score=0.5, booster='gbtree', colsample_bylevel=1, 
    colsample_bynode=1, colsample_bytree=1, gamma=0, gpu_id=-1, 
    importance_type='gain', interaction_constraints='', 
    learning_rate=0.3, max_delta_step=0, max_depth=6, 
    min_child_weight=1, missing=nan, monotone_constraints='()', 
    n_estimators=100, n_jobs=12, num_parallel_tree=1, 
    objective='multi:softprob', random_state=0, reg_alpha=0, 
    reg_lambda=1, scale_pos_weight=None, subsample=1, 
    tree_method='exact', use_label_encoder=False, 
    validate_parameters=1, verbosity=None
)
\end{lstlisting}

Moreover,  OrdinalEncoding is chosen for categorical feature encoding in \tabpfn, which is recommended. For \ours, we set the hyper-parameter $n_e$ of easy samples to the number of shots and $n_h$ of hard samples equal to 50\% of the training data. When LLM-based approaches produce invalid answers that do not fit proper formats, we label them as `no answer'. 
We conduct five-fold cross validation for quantitative comparison. For each dataset and each run, we split the data into training and testing splits with an 80\%/20\% ratio. For our few-shot setting, we randomly sample 16, 32, 64, and 128 instances as the final training sets, along with the full training set for reference. Concerning the large scale of some datasets, we randomly sample 1,000 instances from any test set whose size exceeding 1,000 as the final testing set for resource concerns. 


\section{Extra Experimental Results}
\label{app:extra-exp}

\subsection{Detailed Overall Performance Results}
\label{app:extra-exp-overall}

The results of detailed overall performance with $n=16/32/64/128$ are presented in Table~\ref{tab:main-detailed}. As can be seen, in the very-few shot scenario (\ie $n=16$), both \xgb and \tabpfn could occasionally fail completely. As $n$ increases to 128, these two methods stand out on some datasets. However, LLM-based methods  represented by \ours, are more effective in general. Moreover, \ours obtains the best results in most cases among LLM-based solutions, with only a few exceptions to \tabllm in $n=16/32$. In other words, \ours are better in leveraging the few-shot examples than its competitor \tabllm. This few-shot scenario has great research value for reduced training overhead, as performance can be optimized primarily through effective prompting strategies, saving substantial time in the classification process. Additionally, it has great importance particularly by addressing bias and safety concerns without requiring full data visibility. 

\subsection{Performance Under Full Training Data}
\label{app:extra-exp-fulltrain}

Table~\ref{tab:main-detailed} further presents the results of tested models under full training data, \ie $n=\mbox{all}$. 
Note that \tabpfn is designed for few-shot settings, which encounters runtime errors in this set of tests. 
With an abundant amount of training data, \xgb yields the best performance on six of nine datasets, reclaiming its advantage for general tabular prediction. However, there are still three datasets where \ours outperforms \xgb, indicating the unique values of LLMs for this task.
Finally, \ours is the only LLM-based solution that outperforms \xgb, which we consider a milestone for this topic of study.

\subsection{Comparative Strategy Details}
\label{appendix:comparative}

During the development of \ours, we have explored various implementations of the sampling and grouping strategies. We believe reporting the results of these alternative strategies would also help better understanding our final \ours.
We first summarize the details of these strategies:
\begin{itemize}
    \item Random demonstration (R demos.) only adopts random examples as demonstration, \ie we only add random samples into prompts without any other rules. 
    \item SV demonstration (SV demos.) is another demonstration-only strategy. Specifically, we use the entropy given by an SVC algorithm~\cite{BenHur2002SupportVC} to rank samples and select those easiest as demonstration.
    \item LR rule is a rule-only method without demonstration. It leverages the linear regression algorithm to group training samples with fixed interval and use the rule summarization LLM to summarize rules.
    \item Cocktail combines SV demonstration and  LR rule. To make a fair comparison, we also add the reflective learning strategy to this method.
\end{itemize}

We show the results in Table~\ref{tab:ablation-appendix}. From the table, we can see that the Cocktail approach achieves the best overall score compared to other variants, boasting the highest mistral F1 score across six datasets. This demonstrates the advantage of integrating rules and samples into the prompts, as it allows the LLM to comprehend the data structure at a high level through rules while also providing illustrative examples. 
The improvement is the most evident in the Income dataset, which contains the largest number of data records. 
We also notice that in the Car dataset, using rules-only (LR rule) achieves the best performance. This is likely because the dataset contains four classes, compared to other datasets which typically have only two classes. The presence of more classes might make it easier for the model to derive effective rules that capture the distinctions between categories without requiring additional examples. Overall, these findings highlight the importance of combining rules and samples to enhance the model's performance. By leveraging the strengths of both rules and examples, we can significantly improve the model's ability to understand and process diverse datasets effectively.

\subsection{GPT-3.5 Training Shot Result}
\label{appendix:gpt-3.5-shot}
\begin{figure}
    \centering
    \includegraphics[width=0.9\linewidth]{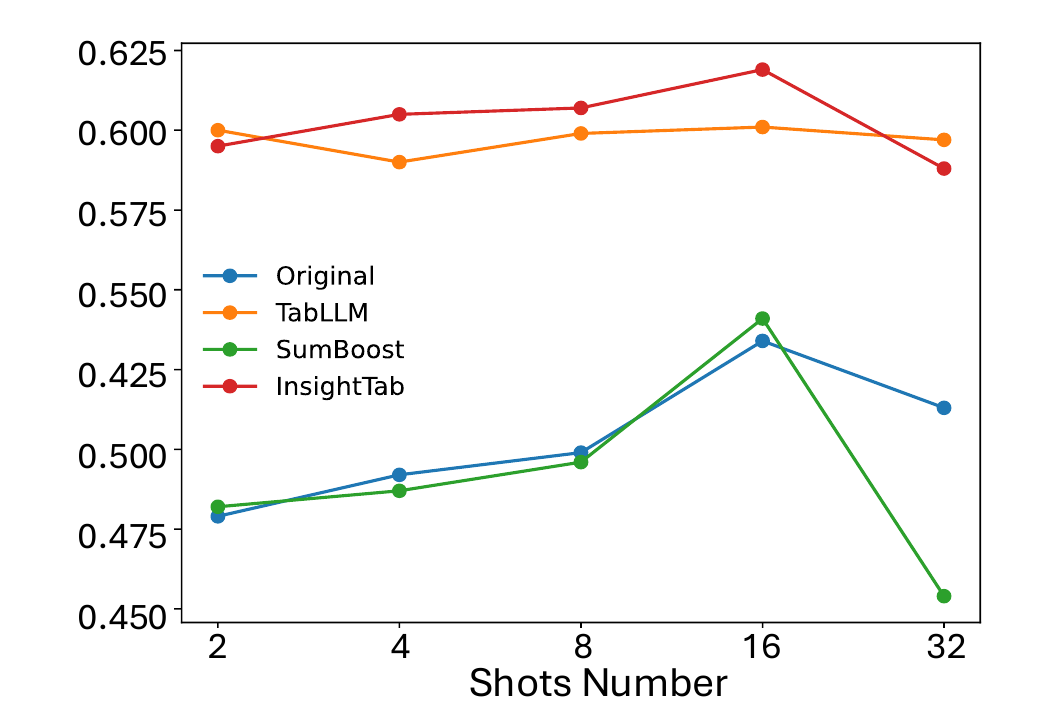}
    \caption{gpt-3.5-turbo performance with 32 training samples
across different number of shots (samples attached in the
prompts). The results are averaged across 9 datasets.}
    \label{fig:gpt-num-shots}
\end{figure}

We show the GPT-3.5-based training shot result in Figure \ref{fig:gpt-num-shots}.

\section{\ours{} Training Cost}
\label{appendix:cost}
We attach the overall cost below in Tables \ref{tab:cost} and \ref{tab:time_cost}. We see our method shows advantage against the most state-of-the-art LLM prompt-based method \summaryboost both in terms of time and budget cost.

\section{LLM Prompts}
\label{appendix:prompts}

\par We finally present prompt templates and prompt examples used in \ours:

\begin{itemize}
    \item \textbf{Prompt template for generating group-level rules} (Fig.~\ref{fig:prompt-group-level-rules}) is used for group-wise rule summarization. Contents in braces are replaced with the meta data of datasets or sample information in application.
    \item \textbf{Prompt template for merging group-level rules} (Fig.~\ref{fig:prompt-summarize-rules}) is used for merging group-wise rules. 
    \item \textbf{Prompt template for tabular classification} (Fig.~\ref{fig:prompt-tabular-prediction-insighttab}) is used for tabular classification based on rules and few-shot samples. 
    \item \textbf{Tabular classification prompt examples} (Figs.~\ref{fig:prompt-example-bank}--\ref{fig:prompt-example-jungle}) are examples of tabular classification prompts on the bank, blood, calhousing, car, creditg, diabetes, heart, income and jungle datasets, respectively. We reserve only one few-shot sample for each prompt due to page size limitation, which is different from the experiment setup. For creditg dataset (Figure \ref{fig:prompt-example-creditg}), we reserve only the first, the second and the last rules due to page size limitation. As mentioned above, additional rules are adaptive in tabular classification prompts.
\end{itemize}

\begin{figure*}
    \small
    \centering
    \begin{tcolorbox}[colback=white,colframe=black,width=\linewidth]
        \{Serialized features of sample 1\} 
        
        \{Question description\}
        
        Answer: \{Label\} 
        
        - - - 

        \ldots 

        - - -

        \{Serialized features of sample $n$\} 
        
        \{Question description\}
        
        Answer: \{Label\} 
        
        - - -
        
        Please distill the key trends that may assist an AI model in making future predictions. Output trends only without any further explanations.
    \end{tcolorbox}
    \caption{Prompt template for generating group-level rules.}
    \label{fig:prompt-group-level-rules}
\end{figure*}

\begin{figure*}
    \small
    \centering
    \begin{tcolorbox}[colback=white,colframe=black,width=\linewidth]
        \{Rules of group 1\}
        
        - - -

        $\ldots$

        - - -

        \{Rules of group n\}
        
        - - -
        
        Tl;dr / Summarize the rules into a small set of non-conflicting and complementary patterns for predicting whether a person earns more than 50000 dollars per year. Output patterns only without any further explanations.
    \end{tcolorbox}
    \caption{Prompt template for merging group-level rules.}
    \label{fig:prompt-summarize-rules}
\end{figure*}

\begin{figure*}
    \small
    \centering
    \begin{tcolorbox}[colback=white,colframe=black,width=\linewidth]
        \{Dataset title\}
        
        \{Dataset description\}
        ~\\
        
        Useful patterns for the task at hand:
        
        \{Summarized rules\}
        ~\\

        Additional patterns for the task summarized from incorrectly classified examples with high entropy:
        
        \{Additional rules\}

        ~\\
        \#\#\#
        ~\\

        [FEW-SHOT EXAMPLES START]
        ~\\
        
        \{Few-shot examples\}
        ~\\
        
        [FEW-SHOT EXAMPLES END]

        ~\\
        \#\#\#
        ~\\

        [CURRENT QUESTION START]

        \{Serialized features of the current question\}
        
        \{Question description\}

        Answer:
    \end{tcolorbox}
    \caption{Prompt template for tabular classification.}
    \label{fig:prompt-tabular-prediction-insighttab}
\end{figure*}

\begin{figure*}
    \small
    \centering
    \begin{tcolorbox}[colback=white,colframe=black,width=\linewidth]
Title: Deposit subscription prediction\\
        This data is related with direct marketing campaigns of a Portuguese banking institution. The marketing campaigns were based on phone calls. The classification goal is to predict if the client will subscribe a term deposit.\\
        ~\\
        Useful patterns for the task at hand:\\
        1. **Previous Campaign Outcome**: Clients with a successful outcome in the previous marketing campaign are more likely to subscribe to a term deposit.\\
2. **Contact Duration**: Longer last contact durations tend to correlate with a higher likelihood of subscription.\\
3. **Housing and Personal Loans**: Clients without housing or personal loans are more likely to subscribe to a term deposit.\\
4. **Number of Contacts in Current Campaign**: A higher number of contacts does not necessarily increase the likelihood of subscription; fewer, more effective contacts might be more beneficial.\\
5. **Communication Type**: While not a strong predictor on its own, effective communication types like cellular phones are commonly used.\\
6. **Yearly Balance**: Clients with higher average yearly balances tend to have a slightly higher likelihood of subscribing, though this is not a strong predictor.\\
7. **Previous Campaign Contact**: Clients not previously contacted in earlier campaigns show varied subscription outcomes, suggesting that this alone is not a strong predictor.\\
8. **Marital Status, Job Type, and Education Level**: These factors do not show a clear trend in influencing the decision to subscribe to a term deposit.\\
9. **Credit in Default**: Clients with no credit default are commonly targeted but this factor alone does not predict subscription.\\
10. **Last Contact Timing (Month and Day)**: No clear trend on how specific months or days impact the likelihood of subscription.\\
        ~\\
        \#\#\#\\
~\\
        {[FEW-SHOT EXAMPLES START]}\\
~\\
The age is 41. The type of job is management. The marital status is married. The education is tertiary. The has credit in default? is no. The average yearly balance, in euros is 144. The has housing loan? is no. The has personal loan? is no. The contact communication type is cellular. The last contact day of the month is 27. The last contact month of year is feb. The last contact duration, in seconds is 123. The number of contacts performed during this campaign and for this client is 3. The number of days that passed by after the client was last contacted from a previous campaign is client was not previously contacted. The number of contacts performed before this campaign and for this client is 0. The outcome of the previous marketing campaign is unknown.\\
~\\
Does this client subscribe to a term deposit?\\
Answer: No\\
~\\
        {[FEW-SHOT EXAMPLES END]}
~\\

        \#\#\#\\
~\\
        {[CURRENT QUESTION START]}
~\\
        The age is 36. The type of job is self-employed. The marital status is married. The education is secondary. The has credit in default? is no. The average yearly balance, in euros is 96. The has housing loan? is no. The has personal loan? is no. The contact communication type is cellular. The last contact day of the month is 27. The last contact month of year is aug. The last contact duration, in seconds is 125. The number of contacts performed during this campaign and for this client is 15. The number of days that passed by after the client was last contacted from a previous campaign is client was not previously contacted. The number of contacts performed before this campaign and for this client is 0. The outcome of the previous marketing campaign is unknown.\\
~\\
Does this client subscribe to a term deposit? Answer the question with either 'Yes' or 'No' (without quotes).\\
Answer: <xxx, No/Yes>
    \end{tcolorbox}
    \caption{A prompt example for Bank dataset.}
    \label{fig:prompt-example-bank}
\end{figure*}

\begin{figure*}
    \small
    \begin{tcolorbox}[colback=white,colframe=black,width=\linewidth]
Title: Blood donation prediction\\
This data is to predict whether a given individual will consent or avoid donating blood.\\
~\\
Useful patterns for the task at hand:\\
1. Lower recency values correlate with a higher likelihood of donating blood.\\
2. Higher frequency of donations increases the likelihood of donating again.\\
3. The total volume of blood donated does not consistently predict donation likelihood.\\
4. Shorter time since the first donation does not consistently predict donation likelihood.\\
5. A combination of low recency and high frequency often predicts a positive donation outcome.\\
~\\
\#\#\#\\
~\\
{[FEW-SHOT EXAMPLES START]}\\
~\\
The Recency - months since last donation is 2. The Frequency - total number of donation is 13. The Monetary - total blood donated in c.c. is 3250. The Time - months since first donation is 53.\\
~\\
Will the person donate blood?\\
Answer: Yes\\
~\\
{[FEW-SHOT EXAMPLES END]}\\
~\\
\#\#\#\\
~\\
{[CURRENT QUESTION START]}\\
~\\
The Recency - months since last donation is 3. The Frequency - total number of donation is 5. The Monetary - total blood donated in c.c. is 1250. The Time - months since first donation is 38.\\
~\\
Will the person donate blood? Answer the question with either 'Yes' or 'No' (without quotes).\\
Answer: <xxx, No/Yes>
    \end{tcolorbox}
    \caption{A prompt example for Blood dataset.}
    \label{fig:prompt-example-blood}
\end{figure*}

\begin{figure*}
    \small
    \begin{tcolorbox}[colback=white,colframe=black,width=\linewidth]
        Title: House value prediction\\
This dataset is collected on the variables using all the block groups in California from the 1990 Census. It computes distances among the centroids of each block group as measured in latitude and longitude, aims to predict whether the house value of district is below or above the median.\\
~\\
Useful patterns for the task at hand:\\
1. Higher median income is associated with higher house block value. \\
2. Lower median age may indicate a more desirable area, potentially increasing property value.\\
3. A higher number of total rooms and bedrooms often correlates with more valuable properties.\\
4. Population size and number of households can suggest demand, but are not definitive indicators of value.\\
5. Geographic location (latitude and longitude) may influence property value based on regional trends.\\
~\\
\#\#\#\\
~\\
{[FEW-SHOT EXAMPLES START]}
~\\
The median income is 3.9097. The median age is 52. The total rooms is 2684. The total bedrooms is 574. The population is 1395. The households is 549. The latitude is 37.76. The longitude is -122.48.\\
~\\
Is this house block valuable?\\
Answer: Yes\\
~\\
- - -\\
~\\
The median income is 3.2344. The median age is 36. The total rooms is 2433. The total bedrooms is 585. The population is 1565. The households is 563. The latitude is 34.08. The longitude is -118.12.\\
~\\
Is this house block valuable?\\
Answer: Yes\\
~\\
{[FEW-SHOT EXAMPLES END]}
~\\

\#\#\#\\
~\\
{[CURRENT QUESTION START]}
~\\
The median income is 3.4659. The median age is 31. The total rooms is 2567. The total bedrooms is 507. The population is 1198. The households is 499. The latitude is 32.78. The longitude is -117.02.\\
~\\
Is this house block valuable? Answer the question with either 'Yes' or 'No' (without quotes).\\
Answer: <xxx, No/Yes>
    \end{tcolorbox}
    \caption{A prompt example for Calhousing dataset.}
    \label{fig:prompt-example-calhousing}
\end{figure*}

\begin{figure*}
    \small
    \begin{tcolorbox}[colback=white,colframe=black,width=\linewidth]
        Title: Car safety prediction\\
        This dataset was derived from a simple hierarchical decision model originally developed for the demonstration of DEX. The goal is to evaluate the safety of cars.\\
~\\
        Useful patterns for the task at hand:\\
        1. Cars accommodating only two persons are generally rated as unacceptable.\\
2. High maintenance costs frequently contribute to an unacceptable rating.\\
3. Safety scores alone do not compensate for negative factors such as low person capacity or high maintenance costs.\\
4. High buying prices combined with very high maintenance costs lead to an unacceptable rating.\\
5. The number of doors and trunk size do not consistently influence the acceptability rating.\\
~\\
        Additional patterns for the task summarized from incorrectly classified examples with high entropy:\\
        1. Low buying price → Unacceptable decision.\\
2. High maintenance costs → Unacceptable decision.\\
3. Low safety score → Unacceptable decision.\\
4. Medium safety score + favorable factors → Acceptable decision.\\
5. More than four persons capacity + negative factors → Unacceptable decision.\\
6. Five or more doors → Favorable but not decisive.\\
7. Big trunk size → Positive factor but not decisive.\\
~\\
        \#\#\#\\
~\\
        {[FEW-SHOT EXAMPLES START]}\\
~\\
The Buying price is medium. The Doors is five or more. The Maintenance costs is very high. The Persons is two. The Safety score is medium. The Trunk size is medium.\\
~\\
How would you rate the decision to buy this car?\\
Answer: Unacceptable\\
~\\
        {[FEW-SHOT EXAMPLES END]}\\
~\\
        \#\#\#\\
~\\
        {[CURRENT QUESTION START]}\\
~\\
        The Buying price is low. The Doors is five or more. The Maintenance costs is medium. The Persons is two. The Safety score is low. The Trunk size is medium.\\
~\\
How would you rate the decision to buy this car? Answer the question with either 'Unacceptable', 'Acceptable', 'Good', or 'Very Good' (without quotes).\\
Answer: <xxx, Unacceptable/Acceptable/Good/Very Good>
    \end{tcolorbox}
    \caption{A prompt example for Car dataset.}
    \label{fig:prompt-example-car}
\end{figure*}

\begin{figure*}
    \small
    \begin{tcolorbox}[colback=white,colframe=black,width=\linewidth]
Title: Credit risk prediction\\
This dataset originates from the UCI Machine Learning Repository. It is used to classify individuals as good or bad credit risks based on various attributes.\\
~\\
Useful patterns for the task at hand:\\
1. **Status of Existing Checking Account**: \\
   - No checking account or a positive balance often leads to approval.\\
   - Higher balances (>= 200 DM) tend to result in denial.\\
2. **Credit History**:\\
   - A history of paying back credits duly often leads to approval.\\
$\ldots$\\
16. **Foreign Worker**:\\
    - Being a foreign worker generally does not prevent credit approval.\\
~\\
\#\#\#\\
~\\
{[FEW-SHOT EXAMPLES START]}\\
~\\
The Status of existing checking account is < 0 DM. The Duration in month is 33. The Credit history  is critical account/ other credits existing (not at this bank). The Purpose is furniture/equipment. The Credit amount is 4281. The Savings account/bonds is 500 <= ... < 1000 DM. The Present employment since is 1 <= ... < 4 years. The Installment rate in percentage of disposable income is 1. The Personal status and sex is female : divorced/separated/married. The Other debtors / guarantors is none. The Present residence since is 4. The Property is car or other, not in attribute 6. The Age in years is 23. The Other installment plans is none. The Housing is own. The Number of existing credits at this bank is 2. The Job is skilled employee / official. The Number of people being liable to provide maintenance for is 1.0. The Telephone is none. The foreign worker is yes.\\
~\\
Does this person receive a credit?\\
Answer: No\\
~\\
{[FEW-SHOT EXAMPLES END]}\\
~\\
\#\#\#\\
~\\
{[CURRENT QUESTION START]}\\
~\\
The Status of existing checking account is < 0 DM. The Duration in month is 9. The Credit history  is existing credits paid back duly till now. The Purpose is car (new). The Credit amount is 654. The Savings account/bonds is ... < 100 DM. The Present employment since is 1 <= ... < 4 years. The Installment rate in percentage of disposable income is 4. The Personal status and sex is male : single. The Other debtors / guarantors is none. The Present residence since is 3. The Property is car or other, not in attribute 6. The Age in years is 28. The Other installment plans is none. The Housing is own. The Number of existing credits at this bank is 1. The Job is unskilled - resident. The Number of people being liable to provide maintenance for is 1.0. The Telephone is none. The foreign worker is yes.\\
~\\
Does this person receive a credit? Answer the question with either 'Yes' or 'No' (without quotes).\\
Answer: <xxx, No/Yes>
    \end{tcolorbox}
    \caption{A prompt example for Creditg dataset.}
    \label{fig:prompt-example-creditg}
\end{figure*}

\begin{figure*}
    \small
    \begin{tcolorbox}[colback=white,colframe=black,width=\linewidth]
Title: Diabetes risk prediction\\
This dataset is originally from the National Institute of Diabetes and Digestive and Kidney Diseases. The objective is to predict based on diagnostic measurements whether a patient has high/low risk of developing diabetes.\\
~\\
Useful patterns for the task at hand:\\
1. **Plasma Glucose Concentration at 2 Hours in GTT**: Values above 140 mg/dL are indicative of diabetes.\\
2. **Body Mass Index (BMI)**: Values over 30 often correlate with diabetes diagnoses.\\
3. **Diabetes Pedigree Function**: Higher values generally correlate with a higher risk of diabetes.\\
4. **Age**: Middle-aged and older adults show a higher prevalence of diabetes.\\
5. **Number of Pregnancies**: Increased number of pregnancies tends to be associated with a higher likelihood of diabetes.\\
6. **2-Hour Serum Insulin**: Higher levels can be indicative of diabetes, especially when combined with other risk factors.\\
7. **Triceps Skin Fold Thickness**: Higher values are often observed in patients with diabetes.\\
8. **Diastolic Blood Pressure**: Readings above 80 mmHg frequently appear in patients diagnosed with diabetes.\\
~\\
\#\#\#\\
~\\
{[FEW-SHOT EXAMPLES START]}\\
~\\
The Age is 52. The Number of times pregnant is 8. The Diastolic blood pressure is 76. The Triceps skin fold thickness is 24. The Plasma glucose concentration at 2 hours in an oral glucose tolerance test (GTT) is 124. The 2-hour serum insulin is 600. The Body mass index is 28.7. The Diabetes pedigree function is 0.687.\\
~\\
Does this patient have diabetes?\\
Answer: Yes\\
~\\
{[FEW-SHOT EXAMPLES END]}\\
~\\
\#\#\#\\
~\\
{[CURRENT QUESTION START]}\\
~\\
The Age is 43. The Number of times pregnant is 7. The Diastolic blood pressure is 80. The Triceps skin fold thickness is 31. The Plasma glucose concentration at 2 hours in an oral glucose tolerance test (GTT) is 109. The 2-hour serum insulin is 0. The Body mass index is 35.9. The Diabetes pedigree function is 1.127.\\
~\\
Does this patient have diabetes? Answer the question with either 'Yes' or 'No' (without quotes).\\
Answer: <xxx, No/Yes>
    \end{tcolorbox}
    \caption{A prompt example for Diabetes dataset.}
    \label{fig:prompt-example-diabetes}
\end{figure*}

\begin{figure*}
    \small
    \begin{tcolorbox}[colback=white,colframe=black,width=\linewidth]
Title: Heart disease prediction\\
This dataset contains 11 features that can be used to predict a possible heart disease.\\
~\\
Useful patterns for the task at hand:\\
1. Presence of exercise-induced angina or asymptomatic chest pain with other risk factors suggests heart disease.\\
2. ST depression induced by exercise relative to rest, especially values of 2.0 or higher, indicates heart disease.\\
3. A flat or downsloping slope of the peak exercise ST segment is indicative of heart disease.\\
4. High fasting blood sugar levels (> 120 mg/dl) are associated with heart disease.\\
5. Lower maximum heart rate achieved during exercise, particularly when combined with other risk factors, suggests heart disease.\\
6. Abnormalities in resting electrocardiogram results (such as ST-T wave abnormality or left ventricular hypertrophy) indicate heart disease, even if chest pain is absent.\\
7. Elevated resting blood pressure and high serum cholesterol levels are risk factors that, when combined with other indicators, suggest heart disease.\\
~\\
Additional patterns for the task summarized from incorrectly classified examples with high entropy:\\
1. Older age increases the likelihood of heart disease.\\
2. Asymptomatic cases are more likely to indicate heart disease compared to non-anginal pain.\\
3. Elevated serum cholesterol levels are indicative of heart disease.\\
4. Elevated fasting blood sugar (> 120 mg/dl) may suggest heart disease.\\
5. Abnormal resting electrocardiogram results indicate a higher risk of heart disease.\\
6. Presence of exercise-induced angina is a strong indicator of heart disease.\\
7. Higher ST depression induced by exercise is associated with heart disease.\\
8. Flat slopes of the peak exercise ST segment may indicate heart disease.\\
~\\
\#\#\#\\
~\\
{[FEW-SHOT EXAMPLES START]}\\
~\\
The Age of the patient is 46. The Sex of the patient is male. The Chest pain type is asymptomatic. The Resting blood pressure is 120. The Serum cholesterol is 231. The Fasting blood sugar > 120 mg/dl is no. The Resting electrocardiogram results is normal. The Maximum heart rate achieved is 115. The Exercise-induced angina is yes. The ST depression induced by exercise relative to rest is 0.0. The Slope of the peak exercise ST segment is flat.\\
~\\
Does the coronary angiography of this patient show a heart disease?\\
Answer: Yes\\
~\\
{[FEW-SHOT EXAMPLES END]}\\
~\\
\#\#\#\\
~\\
{[CURRENT QUESTION START]}\\
~\\
The Age of the patient is 59. The Sex of the patient is male. The Chest pain type is asymptomatic. The Resting blood pressure is 170. The Serum cholesterol is 326. The Fasting blood sugar > 120 mg/dl is no. The Resting electrocardiogram results is probable or definite left ventricular hypertrophy. The Maximum heart rate achieved is 140. The Exercise-induced angina is yes. The ST depression induced by exercise relative to rest is 3.4. The Slope of the peak exercise ST segment is downsloping.\\
~\\
Does the coronary angiography of this patient show a heart disease? Answer the question with either 'Yes' or 'No' (without quotes).
Answer: <xxx, No/Yes>
    \end{tcolorbox}
    \caption{A prompt example for Heart dataset.}
    \label{fig:prompt-example-heart}
\end{figure*}

\begin{figure*}
    \small
    \begin{tcolorbox}[colback=white,colframe=black,width=\linewidth]
Title: Income prediction\\
This dataset predicts whether each person has an annual income over \$50,000 based on some information about the person.\\
~\\
Useful patterns for the task at hand:\\
1. Higher education levels, particularly master's and doctoral degrees, are associated with higher earnings.\\
2. Occupations in management, execution, professional specialties, and protective services (especially in local government) are linked to higher earnings.\\
3. Employment in government sectors (state and local) and ownership of incorporated businesses tend to correlate with higher earnings.\\
4. Working 40 hours per week or more is generally associated with higher earnings, particularly in professional, managerial, or government roles.\\
5. Marital status as married and relation as husband often correlates with higher earnings.\\
6. Capital gains can indicate higher earnings, but capital losses do not necessarily indicate lower earnings.\\
7. Native-born U.S. citizens do not show a clear trend affecting earnings compared to other countries based on the provided data.\\
8. Younger individuals, especially those under 25, are less likely to earn more than \$50,000, regardless of other factors.\\
~\\
Additional patterns for the task summarized from incorrectly classified examples with high entropy:\\
1. Higher education levels (master's degree or higher) are associated with earning more than \$50,000 per year.\\
2. Occupations in professional specialties are likely to earn more than \$50,000, while occupations in agriculture or with lower education levels typically do not.\\
3. Working higher hours (60 or more) in certain sectors can lead to earnings above \$50,000.\\
4. Local government positions' incomes vary based on occupation and education level, potentially leading to earnings above \$50,000.\\
5. Capital gains and losses do not directly affect annual earnings above or below \$50,000.\\
~\\
\#\#\#\\
~\\
{[FEW-SHOT EXAMPLES START]}\\
~\\
The Age is 23. The Race is White. The Sex is Male. The  Marital status is married. The Relation to head of the household is Husband. The Native country is United States. The Occupation is sales sector. The Work class is private sector employee. The Capital gain last year is 0. The Capital loss last year is 0. The Education is high school graduate. The Work hours per week is 40.\\
~\\
Does this person earn more than 50000 dollars per year?\\
Answer: Yes\\
~\\
{[FEW-SHOT EXAMPLES END]}\\
~\\
\#\#\#\\
~\\
{[CURRENT QUESTION START]}\\
~\\
The Age is 32. The Race is White. The Sex is Female. The  Marital status is separated. The Relation to head of the household is Unmarried. The Native country is United States. The Occupation is sales sector. The Work class is private sector employee. The Capital gain last year is 0. The Capital loss last year is 0. The Education is finished 8th class. The Work hours per week is 35.\\
~\\
Does this person earn more than 50000 dollars per year? Answer the question with either 'Yes' or 'No' (without quotes).\\
Answer: <xxx, No/Yes>
    \end{tcolorbox}
    \caption{A prompt example for Income dataset.}
    \label{fig:prompt-example-income}
\end{figure*}

\begin{figure*}
    \small
    \begin{tcolorbox}[colback=white,colframe=black,width=\linewidth]
Title: Game winning prediction\\
        This dataset predicts whether the white player will win based on some game information.\\
~\\
        Useful patterns for the task at hand:\\
        1. White wins if the white piece has higher strength than the black piece, unless the positional difference is too great.\\
2. White does not win when both pieces have equal strength.\\
3. Proximity of the white piece to the opponent's den (lower ranks) increases the likelihood of a win for white.\\
4. The file position (horizontal alignment) of the pieces does not significantly influence the outcome.\\
5. White does not win when the white piece has the lowest strength (0), regardless of the black piece's position or strength.\\
6. Strategic positioning, such as cornering the opponent's piece, can increase the likelihood of winning for the stronger piece.\\
~\\
        Additional patterns for the task summarized from incorrectly classified examples with high entropy:\\
        1. Higher piece strength generally favors the player with the stronger piece.\\
2. Positioning on the board (file and rank) can influence the outcome, particularly in endgames.\\
3. The ability to control key squares or escape routes is crucial for determining the winner.\\
4. When piece strengths are equal, positioning becomes more significant in deciding the outcome.\\
5. A piece with greater strength positioned to attack is likely to lead to a win.\\
~\\
        \#\#\#\\
~\\
        {[FEW-SHOT EXAMPLES START]}\\
~\\
The white piece strength is 4. The white piece file is 1. The white piece rank is 7. The black piece strength is 0. The black piece file is 4. The black piece rank is 8.\\
~\\
Does the white player win this two pieces endgame of Jungle Chess?\\
Answer: Yes\\
~\\
        {[FEW-SHOT EXAMPLES END]}\\
~\\
        \#\#\#\\
~\\
        {[CURRENT QUESTION START]}\\
~\\
        The white piece strength is 7. The white piece file is 2. The white piece rank is 0. The black piece strength is 0. The black piece file is 1. The black piece rank is 4.\\
~\\
Does the white player win this two pieces endgame of Jungle Chess? Answer the question with either 'Yes' or 'No' (without quotes).\\
Answer: <xxx, No/Yes>
    \end{tcolorbox}
    \caption{A prompt example for Jungle dataset.}
    \label{fig:prompt-example-jungle}
\end{figure*}

\end{appendix}
\end{document}